\documentclass[11pt]{article}

\let\ACLbibliographystyle\bibliographystyle
\renewcommand{\bibliographystyle}[1]{\ACLbibliographystyle{halo_natbib}}
\usepackage[preprint]{acl}

\usepackage{times}
\usepackage{latexsym}
\usepackage{amssymb}
\usepackage{fontawesome5}

\usepackage[T1]{fontenc}

\usepackage[utf8]{inputenc}

\usepackage{microtype}

\usepackage{inconsolata}

\usepackage{graphicx}
\usepackage{tikz}
\usepackage{booktabs}
\usepackage{float}
\usepackage{array}
\usepackage{colortbl}
\usepackage{tcolorbox}
\tcbuselibrary{breakable}
\usepackage{fvextra}
\definecolor{promptfill}{RGB}{244,243,255}
\definecolor{promptframe}{RGB}{76,87,186}
\newenvironment{haloprompt}{%
  \VerbatimEnvironment
  \begin{tcolorbox}[breakable,colback=promptfill,colframe=promptframe,
    boxrule=0.45pt,arc=0pt,boxsep=0pt,left=5pt,right=5pt,top=5pt,bottom=5pt,
    before skip=4pt,after skip=5pt]
  \begin{Verbatim}[fontsize=\footnotesize,breaklines=true,breakanywhere=true,
    breaksymbolleft={},breaksymbolright={},breakindent=0pt]%
}{%
  \end{Verbatim}
  \end{tcolorbox}%
}
\usepackage{stfloats}
\definecolor{findingframe}{RGB}{76,87,186}
\definecolor{findingtext}{RGB}{51,108,151}
\definecolor{findingfill}{RGB}{249,250,255}
\newtcolorbox{finding}{colback=findingfill,colframe=findingframe,
  boxrule=0.45pt,arc=2pt,boxsep=0pt,left=4pt,right=4pt,top=3pt,bottom=3pt,
  before skip=3pt,after skip=3pt,fontupper=\bfseries\color{findingtext}}
\newlength{\halomodelwidth}
\newlength{\halonumwidth}
\definecolor{tableband}{RGB}{237,239,242}

\definecolor{haloblue}{HTML}{386B9D}
\definecolor{haloorange}{HTML}{C98446}
\DeclareRobustCommand{\haloicon}{%
  \tikz[baseline=-0.1ex,x=1em,y=1em]{%
    \path[use as bounding box] (-0.04,-0.02) rectangle (1.14,0.90);
    \begin{scope}[rotate around={14:(0.52,0.40)}]
      \path[fill=haloorange,even odd rule]
        (0.52,0.40) ellipse[x radius=0.53,y radius=0.20]
        (0.52,0.40) ellipse[x radius=0.39,y radius=0.105];
    \end{scope}
    \path[fill=haloorange]
      (1.00,0.64) -- (1.03,0.72) -- (1.11,0.75) -- (1.03,0.78) --
      (1.00,0.86) -- (0.97,0.78) -- (0.89,0.75) -- (0.97,0.72) -- cycle;
  }%
}
\title{\texorpdfstring{%
  \haloicon\hspace{0.20em}%
  \textcolor{haloblue}{H}\textcolor{haloorange}{A}%
  \textcolor{haloblue}{L}\textcolor{haloorange}{O}: %
  \textcolor{haloblue}{H}uman-Grounded Gr\textcolor{haloorange}{A}nular
  Eva\textcolor{haloblue}{L}uation of Pr\textcolor{haloorange}{O}active Agents%
}{HALO: Human-Grounded GrAnular EvaLuation of PrOactive Agents}}

\author{
  \textbf{Wenqi Pei\textsuperscript{1}},\quad
  \textbf{Xinyu Wang\textsuperscript{2}},\quad
  \textbf{Jinrui Liu\textsuperscript{2}},\quad
  \textbf{Cynthia Zhuge\textsuperscript{2}},\quad
  \textbf{Jie Sheng\textsuperscript{2}} \\[0.3em]
  \textbf{Zihan Wang\textsuperscript{2}},\quad
  \textbf{Tenghao Huang\textsuperscript{3}},\quad
  \textbf{Muhao Chen\textsuperscript{1}} \\[0.65em]
  \textsuperscript{1}University of California, Davis
  \quad \textsuperscript{2}2077AI
  \quad \textsuperscript{3}University of Southern California
}

\begin{document}
\maketitle
\begin{abstract}
Proactive agents must decide whether, when, and how far to assist without an explicit request.
Existing evaluations often rely on automated scores or binary verdicts, offering limited human diagnosis of evidence use and intervention quality.
To address these gaps, we introduce \textbf{HALO}\footnote{\faGithub\hspace{0.3em}\href{https://github.com/CedricPei/HALO/}{\nolinkurl{github.com/CedricPei/HALO}}}, \textbf{H}uman-Grounded Gr\textbf{A}nular Eva\textbf{L}uation of Pr\textbf{O}active Agents.
\textbf{HALO-Arena} covers 4 proactive need types, Dependency, Blockage, Risk, and Opportunity, across 10 everyday domains and 50 personal environments.
Scenarios require agents to retrieve evidence from structured databases or unstructured documents before judging whether assistance is warranted and choosing its scope.
On HALO-Arena, we construct \textbf{HALO-8K}, a dataset of 8,000 responses from 10 models across 400 scenarios, annotated by 20 experts for action level, acceptability verdicts, and human diagnosis.
For rejected responses, human diagnosis covers timing, intensity, and other issues.
Recorded reasoning tokens and retrieval rounds measure the effort behind these responses.
The best proprietary and open-source models achieve 72.88\% and 56.63\% acceptability, respectively.
Our central finding is that over-proactivity is the dominant observed failure mode, with Premature timing or Excessive intensity accounting for 61.20\% of rejected responses.
Further analyses yield complementary findings on the balance between intervention and restraint, evidence acquisition, and the calibration of timing and intensity.
\end{abstract}

\section{Introduction}
\label{sec:introduction}

Proactive agents infer user needs from observed activity and initiate assistance without an explicit request \citep{lu2024proactiveagent}.
They have attracted growing attention for their potential to anticipate problems and reduce user effort \citep{tang2026proagentbench,oh2026proactivitygym}.
They can identify missing prerequisites, anticipate risks, and prepare useful next steps within the user's authority.
For example, \citet{lu2024proactiveagent} train agents to propose tasks from observed user activities and report a 66.47\% F1 score for proactive assistance.
Useful initiative requires deciding whether, when, and how far to help, while recognizing when restraint is warranted.

\begin{figure*}[t]
  \centering
  \includegraphics[width=\textwidth]{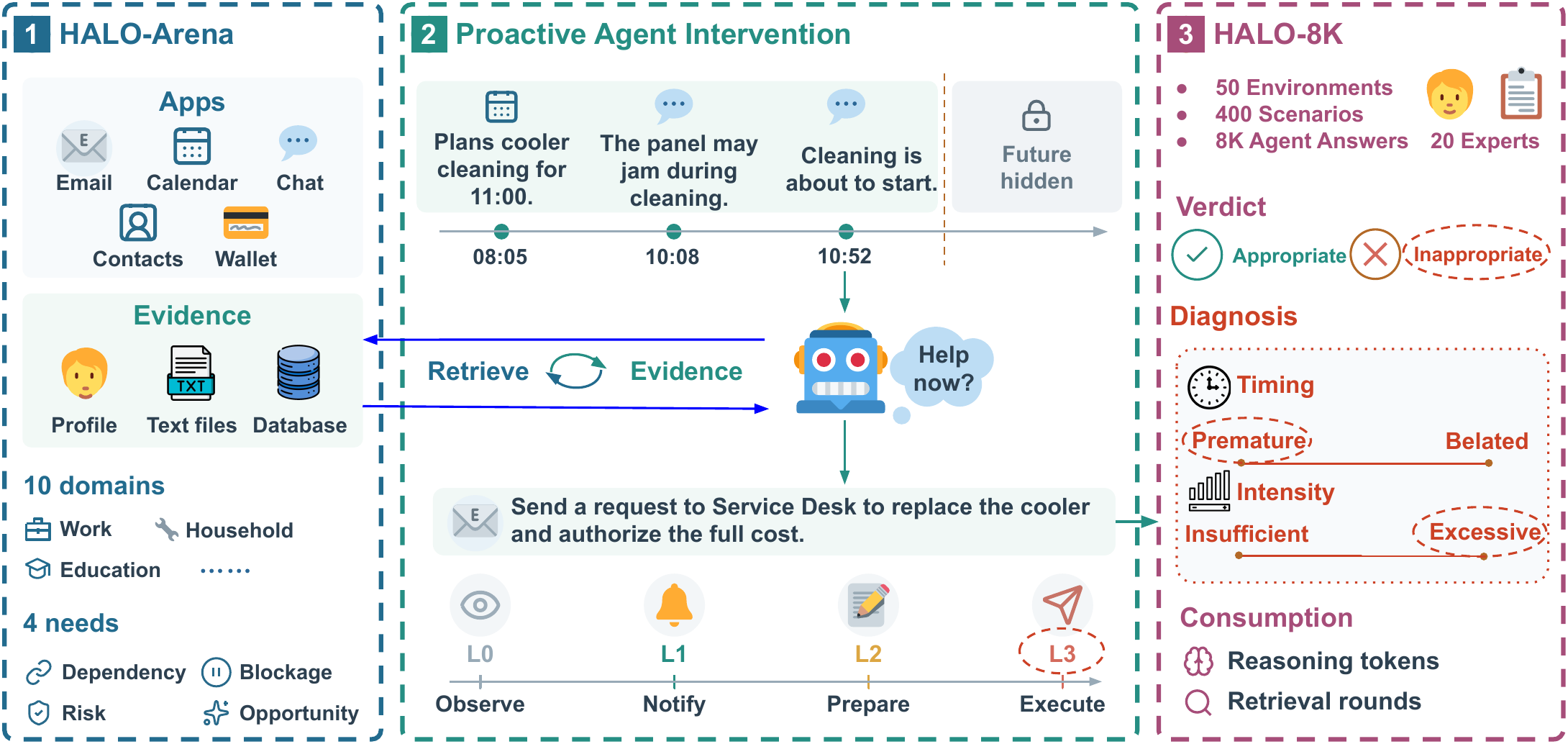}
  \caption{Overview of HALO. HALO-Arena supplies personal apps and local evidence. Agents retrieve evidence and decide whether, when, and how far to intervene. HALO-8K records human verdicts, timing and intensity diagnosis, and resource consumption. The circled choices illustrate a premature and excessive L3 intervention.}
  \label{fig:overview}
\end{figure*}

A calendar entry for an upcoming trip may prompt checking a database for needed supplies and a file for purchase authorization.
These facts can justify a reminder without permitting a purchase.
Retrieved evidence must guide the intervention's timing, inform its content, and establish the authorized scope.

Benchmarks differ in how agents obtain evidence for proactive decisions.
Some provide user activity directly \citep{lu2024proactiveagent,tang2026proagentbench}.
Such observations may signal a need without establishing a remedy or permission to act.
PROBE supports SQL and semantic document retrieval \citep{pasternak2025probe}, while interactive benchmarks support additional tool use \citep{oh2026proactivitygym,zhang2026pibench}.
Evaluating these capabilities requires linking retrieved evidence to human verdicts on the resulting interventions.

Detecting a relevant need does not by itself make an intervention acceptable. An intervention may still fail because it is mistimed, insufficient, excessive, irrelevant, or unsupported by evidence. Existing benchmarks primarily rely on scalable automated metrics: ProAgentBench evaluates intervention prediction and response similarity \citep{tang2026proagentbench}, while ProactiveBench uses a reward model trained on human acceptance labels \citep{lu2024proactiveagent}. Human studies have examined how users perceive proactive assistance \citep{oh2026proactivitygym}, but existing benchmarks provide limited human diagnosis of why individual model responses are accepted or rejected.

To address these gaps, we introduce \textbf{HALO}, \textbf{H}uman-Grounded Gr\textbf{A}nu\-lar Eva\textbf{L}u\-a\-tion of Pr\textbf{O}ac\-tive Agents, with 2 components.
HALO combines structured database querying and unstructured file retrieval with fine-grained human evaluation.
\textbf{HALO-Arena} models 4 proactive need types: Dependency, Blockage, Risk, and Opportunity.
It spans 10 everyday domains and 50 personal environments.
Operational applications expose evolving events, while separate databases and text files provide structured and unstructured evidence.
Addressing a need requires the agent to retrieve relevant evidence and then judge whether assistance is warranted, when to act, and how far to intervene.
At each evaluation checkpoint, agents receive an observed event prefix and may retrieve evidence over multiple rounds before returning 1 executable response or no action.

On HALO-Arena, we construct \textbf{HALO-8K}, comprising 8,000 responses from 10 models across 400 scenarios, annotated by 20 experts. Annotators assign action levels and acceptability verdicts, with human diagnosis of rejections covering timing, intensity, and other issues such as unsupported evidence, irrelevance, and factual contradiction. Verdicts on no-action responses distinguish erroneous Inaction from warranted restraint. Reasoning tokens and retrieval rounds quantify agent effort.

Our central finding is that \textbf{over-proactivity is the dominant observed failure mode}: Premature timing or Excessive intensity accounts for 61.20\% of rejected responses. Proprietary and open-source models peak at 72.88\% and 56.63\% acceptability. Further analysis yields 3 complementary findings. \textbf{(1) Restraint contributes to acceptability:} gains in accepted action can be outweighed by losses in warranted restraint. \textbf{(2) Evidence acquisition accompanies better help:} among reminders, local content access is associated with higher acceptability and lower Insufficient diagnosis rates. \textbf{(3) Timing alone is not enough:} among well-timed rejections, reminders tend to be insufficient and executions excessive. These findings support granular evaluation of proactive assistance.

Our contributions are as follows:
\begin{list}{\labelitemi}{ \setlength{\leftmargin}{12pt} \setlength{\labelwidth}{8pt} \setlength{\labelsep}{4pt} \setlength{\itemindent}{0pt} \setlength{\topsep}{1pt} \setlength{\partopsep}{0pt} \setlength{\itemsep}{1pt} \setlength{\parsep}{0pt} }
\item We introduce \textbf{HALO-Arena}, an executable benchmark spanning 4 proactive need types and 10 everyday domains, with evidence retrieval informing proactive decisions.
\item We construct \textbf{HALO-8K}: 8,000 responses across 400 scenarios, annotated by experts for action level, verdicts, and human diagnosis.
\item We evaluate 10 models, revealing tradeoffs among intervention, restraint, timing, intensity, and retrieval effort.
\end{list}

\section{Related Work}
\label{sec:related_work}

\subsection{Proactive Agents}

Early agents predict useful tasks from observed activity and refine proposals using human acceptance verdicts \citep{lu2024proactiveagent}.
ContextAgent extends perception to wearable video and audio, combining sensory and persona context for tool-assisted services \citep{yang2025contextagent}.

Recent work also examines timing and scope.
ProActor models actionable windows for timing-aware scheduling \citep{ding2026proactor}.
Proactivity-Gym connects task capability with temporal allocation and user trust \citep{oh2026proactivitygym}.

\subsection{Benchmarks for Proactive Agents}

Evaluation has expanded from executing explicit requests to identifying unstated needs.
WebArena and $\tau$-bench test request-driven task execution \citep{zhou2023webarena,yao2024taubench}.
ProactiveBench evaluates proposals from activity traces \citep{lu2024proactiveagent}, while ContextAgentBench adds multimodal context and tool calling \citep{yang2025contextagent}.
PROBE evaluates bottleneck discovery and action selection from document stores, while its agent baselines use SQL and semantic retrieval \citep{pasternak2025probe}.

Recent benchmarks examine proactive assistance in interactive settings.
ProAgentBench evaluates intervention decisions and generated assistance in real-world computer activity \citep{tang2026proagentbench}.
FingerTip~20K covers proactive intent suggestions and personalized Android execution \citep{yang2026fingertip}.
PARE-Bench introduces stateful apps and active user simulation \citep{nathani2026pare}, while $\pi$-Bench evaluates hidden-intent resolution and task completion across persistent sessions \citep{zhang2026pibench}.

Complementary evaluations focus on intervention calibration.
AgentAbstain tests action and restraint under explicit requests \citep{liu2026agentabstain}.
ProEvent evaluates response timing and timetable updates in ongoing chats \citep{li2026proevent}.
Proactivity-Gym measures temporal allocation, intervention depth, and trust in multi-day simulations and includes a human study \citep{oh2026proactivitygym}.

HALO combines executable apps, structured database queries, and unstructured file retrieval with human verdicts, granular human diagnosis, and recorded resource consumption.
Appendix~\ref{sec:benchmark_comparison} compares benchmark environments and evaluation.

\begingroup
\makeatletter
\renewcommand\paragraph{\@startsection{paragraph}{4}{\z@}{1.2ex plus .2ex minus .2ex}{-1em}{\normalfont\normalsize\bfseries}}
\makeatother
\section{HALO}
\label{sec:benchmark}

\subsection{HALO-Arena}
\label{sec:task}

\subsubsection{Overview}
\label{sec:arena_overview}

\paragraph{Applications.}
HALO-Arena combines email, calendar, chat, contacts, and wallet apps with SQLite databases and text files.
Agents use 29 interaction commands for evidence retrieval and 19 action commands for attention, preparation, and execution.
Appendix~\ref{sec:app_commands} lists the apps and commands by category.

\paragraph{Domains.}
The 10 domains are education and learning, work and professional life, home and household management, family and caregiving, health and wellbeing, travel and mobility, personal finance and consumer life, community and civic life, social and cultural life, and recreation (sports and hobbies).
Each domain has 5 instances, each representing a different person, yielding 50 distinct personal environments.

\paragraph{Proactive needs.}
We consider 4 need types.
\underline{\textit{Dependency}} is a missing prerequisite for a future activity.
\underline{\textit{Blockage}} occurs when an ongoing activity cannot continue.
\underline{\textit{Risk}} is a possible adverse outcome that has not yet occurred.
\underline{\textit{Opportunity}} is an optional benefit whose omission causes no failure or loss.
Each environment contains 2 scenarios per need type, yielding 8 scenarios.

As shown in Figure~\ref{fig:overview}, HALO-Arena provides executable environments for agent actions, with 400 designed scenarios.

\begin{table*}[t]
  \centering
  \small
  \definecolor{rankbest}{RGB}{218,238,214}
  \definecolor{ranksecond}{RGB}{219,232,247}
  \definecolor{rankthird}{RGB}{252,236,209}
  \definecolor{modelband}{RGB}{235,235,235}
  \caption{Model performance and resource consumption. Action shares and error rates are percentages of all responses. Inact., Prem., Bel., Insuff., and Excess. denote Inaction, Premature, Belated, Insufficient, and Excessive. Shading marks \colorbox{rankbest}{best}, \colorbox{ranksecond}{second}, and \colorbox{rankthird}{third}.}
  \label{tab:main_results}
  \setlength{\tabcolsep}{1pt}
  \settowidth{\halomodelwidth}{DeepSeekV4-Pro}
  \setlength{\halonumwidth}{\dimexpr(\textwidth-\halomodelwidth-4pt-22\tabcolsep)/12\relax}
  \renewcommand{\arraystretch}{1.20}
  \begin{tabular}{@{}l@{\hspace{4pt}}*{12}{>{\centering\arraybackslash}m{\halonumwidth}}@{}}
    \toprule
    \raisebox{-1ex}{Model} & \multicolumn{4}{c}{\footnotesize Action Levels} & \makebox[\linewidth][c]{\raisebox{-1ex}{\footnotesize Inact.$\downarrow$}} & \multicolumn{2}{c}{\footnotesize Timing Errors$\downarrow$} & \multicolumn{2}{c}{\footnotesize Intensity Errors$\downarrow$} & \multicolumn{2}{c}{\footnotesize Consumption$\downarrow$} & \makebox[\linewidth][c]{\raisebox{-1ex}{\footnotesize Acc (\%)$\uparrow$}} \\[-2.5pt]
    \cmidrule(lr){2-5}\cmidrule(lr){7-8}\cmidrule(lr){9-10}\cmidrule(lr){11-12}
    & {\footnotesize L0} & {\footnotesize L1} & {\footnotesize L2} & {\footnotesize L3} & & {\footnotesize Prem.} & {\footnotesize Bel.} & {\footnotesize Insuff.} & {\footnotesize Excess.} & \makebox[\linewidth][c]{\footnotesize Reas. (k)} & {\footnotesize Rounds} & \\
    \specialrule{0.4pt}{2pt}{0pt}
    \rowcolor{modelband}\multicolumn{13}{c}{\raisebox{-2pt}{\normalsize\bfseries\itshape Proprietary Models}} \\[3pt]
    \specialrule{0.4pt}{0pt}{2pt}
    Gemini3.8-Flash & 30.13 & 57.50 & 2.13 & 10.25 & 7.38 & \cellcolor{rankbest}10.50 & \cellcolor{rankbest}3.75 & \cellcolor{ranksecond}2.63 & \cellcolor{rankbest}7.00 & 14.99 & 4.19 & \cellcolor{rankbest}72.88 \\
    Claude-Opus5.5 & 31.75 & 48.00 & 5.13 & 15.13 & 6.75 & \cellcolor{ranksecond}10.88 & \cellcolor{ranksecond}3.88 & \cellcolor{rankbest}2.00 & \cellcolor{ranksecond}8.38 & 3.48 & 2.27 & \cellcolor{ranksecond}69.75 \\
    Grok4.7 & 3.63 & 76.50 & 2.00 & 17.88 & 0.88 & \cellcolor{rankthird}21.38 & 6.88 & \cellcolor{rankthird}3.25 & 14.63 & 5.04 & 2.70 & \cellcolor{rankthird}65.25 \\
    Gemini3.1-Pro & 3.88 & 58.63 & 7.63 & 29.88 & \cellcolor{ranksecond}0.13 & 27.63 & \cellcolor{rankthird}6.13 & 4.63 & 19.25 & 3.80 & 1.85 & 55.75 \\
    GPT5.6-Luna & 2.50 & 87.00 & 3.25 & 7.25 & \cellcolor{ranksecond}0.13 & 26.25 & 9.38 & 10.63 & 14.75 & \cellcolor{rankbest}0.36 & 1.77 & 51.75 \\
    \textbf{Avg.} & 14.38 & 65.53 & 4.03 & 16.08 & 3.05 & 19.33 & 6.00 & 4.63 & 12.80 & 5.53 & 2.55 & 63.08 \\
    \specialrule{0.4pt}{2pt}{0pt}
    \rowcolor{modelband}\multicolumn{13}{c}{\raisebox{-2pt}{\normalsize\bfseries\itshape Open-source Models}} \\[3pt]
    \specialrule{0.4pt}{0pt}{2pt}
    DeepSeekV4-Pro & 2.63 & 79.50 & 2.25 & 15.63 & \cellcolor{ranksecond}0.13 & 24.25 & 8.25 & 7.50 & 16.13 & 5.42 & \cellcolor{rankthird}1.39 & 56.63 \\
    Qwen3.8-27B & 0.38 & 68.50 & 3.75 & 27.38 & \cellcolor{rankbest}0.00 & 25.38 & 7.88 & 5.50 & 19.75 & 9.86 & 1.54 & 55.13 \\
    GLM5.3-Flash & 4.88 & 73.63 & 2.88 & 18.63 & \cellcolor{rankthird}0.75 & 23.88 & 7.75 & 5.75 & 18.25 & \cellcolor{rankthird}1.48 & 1.82 & 54.13 \\
    Qwen3.5-9B & 20.63 & 69.00 & 4.38 & 6.00 & 9.63 & 22.50 & 7.75 & 13.00 & \cellcolor{rankthird}11.75 & 6.37 & \cellcolor{ranksecond}0.96 & 41.50 \\
    GPT-OSS-20B & 5.00 & 73.75 & 8.38 & 12.88 & 1.50 & 27.75 & 13.75 & 17.38 & 17.25 & \cellcolor{ranksecond}0.75 & \cellcolor{rankbest}0.37 & 30.38 \\
    \textbf{Avg.} & 6.70 & 72.88 & 4.33 & 16.10 & 2.40 & 24.75 & 9.08 & 9.83 & 16.63 & 4.77 & 1.21 & 47.55 \\
    \bottomrule
  \end{tabular}
\end{table*}

\subsubsection{Construction}
\label{sec:generation}

\begin{figure}[t]
  \centering
  \captionsetup{skip=5pt}
  \includegraphics[width=\columnwidth]{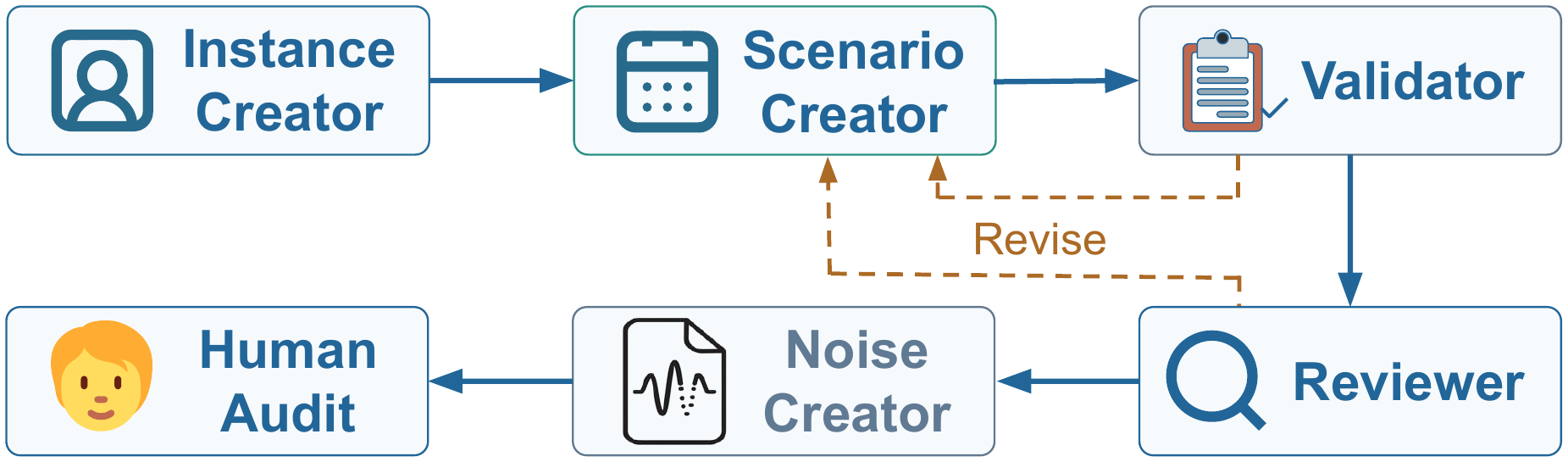}
  \caption{HALO-Arena construction. Validator and Reviewer return revision feedback before background records and human audit finalize the dataset.}
  \label{fig:construction}
  \vspace{-8pt}
\end{figure}

\paragraph{Creators.}
We use GPT5.6-Sol to construct HALO-Arena through the pipeline shown in Figure~\ref{fig:construction}.
Instance Creator defines personal responsibilities, authority, and application and storage structures.
Scenario Creator populates these structures to form an initial state $S_0$ and generates $T$ timestamped events $(e_1,\ldots,e_T)$ for each scenario, where $e_t$ denotes event $t$.
Trajectories span at least 3 operational apps and include prefixes where intervention would be premature or belated, with local evidence for that decision but no explicit help requests or agent interventions.
After all 8 scenarios pass review, Noise Creator adds settled background records to $S_0$ to broaden retrieval.

\paragraph{Validator and Reviewer.}
Validator deterministically checks structure, command signatures, state consistency, and isolated execution.
Reviewer assesses causal support, need identity, timing, retrieval necessity, and proportionate assistance.
Scenario Creator revises failed scenarios.

\paragraph{Human Audit.}
\label{sec:audit}

2 reviewers audit each scenario for causal support, timing, evidence, and feasible assistance within authority.
A third reviewer adjudicates cases with exactly 1 initial approval.
Scenarios that do not pass review are revised before model-based acceptance checks.
Detailed procedures and results appear in Appendix~\ref{sec:construction_details}.

\subsection{HALO-8K}
\label{sec:halo_8k}
\label{sec:human_eval}
\label{sec:protocol}
\label{sec:evaluation}

Built on HALO-Arena, HALO-8K contains 8,000 human-annotated responses from 10 models across 2,160 checkpoints.

\begin{table*}[htb]
  \centering
  \small
  \definecolor{rankbest}{RGB}{218,238,214}
  \definecolor{ranksecond}{RGB}{219,232,247}
  \definecolor{rankthird}{RGB}{252,236,209}
  \definecolor{contextband}{RGB}{235,235,235}
  \caption{Results by proactive need type and domain. All values are percentages of category responses. Error abbreviations follow Table~\ref{tab:main_results}. Shading marks \colorbox{rankbest}{best}, \colorbox{ranksecond}{second}, and \colorbox{rankthird}{third} values within each group.}
  \label{tab:context_results}
  \setlength{\tabcolsep}{1pt}
  \settowidth{\halomodelwidth}{Dependency}
  \setlength{\halonumwidth}{\dimexpr(\textwidth-\halomodelwidth-4pt-18\tabcolsep)/10\relax}
  \renewcommand{\arraystretch}{1.20}
  \begin{tabular}{@{}l@{\hspace{4pt}}*{10}{>{\centering\arraybackslash}m{\halonumwidth}}@{}}
    \toprule
    \raisebox{-1ex}{Category} & \multicolumn{4}{c}{\footnotesize Action Levels} & \makebox[\linewidth][c]{\raisebox{-1ex}{\footnotesize Inact.$\downarrow$}} & \multicolumn{2}{c}{\footnotesize Timing Errors$\downarrow$} & \multicolumn{2}{c}{\footnotesize Intensity Errors$\downarrow$} & \makebox[\linewidth][c]{\raisebox{-1ex}{\footnotesize Acc (\%)$\uparrow$}} \\[-2.5pt]
    \cmidrule(lr){2-5}\cmidrule(lr){7-8}\cmidrule(lr){9-10}
    & {\footnotesize L0} & {\footnotesize L1} & {\footnotesize L2} & {\footnotesize L3} & & {\footnotesize Prem.} & {\footnotesize Bel.} & {\footnotesize Insuff.} & {\footnotesize Excess.} & \\
    \specialrule{0.4pt}{2pt}{0pt}
    \rowcolor{contextband}\multicolumn{11}{c}{\raisebox{-2pt}{\normalsize\bfseries\itshape Proactive Need Types}} \\[3pt]
    \specialrule{0.4pt}{0pt}{2pt}
    Blockage & 12.05 & 66.35 & 4.35 & 17.25 & 3.15 & \cellcolor{ranksecond}21.50 & \cellcolor{rankthird}7.95 & 8.70 & \cellcolor{ranksecond}13.85 & 54.35 \\
    Dependency & 7.20 & 71.45 & 4.95 & 16.40 & \cellcolor{rankbest}1.95 & \cellcolor{rankbest}21.05 & \cellcolor{ranksecond}7.80 & \cellcolor{ranksecond}7.10 & 16.05 & \cellcolor{rankthird}55.05 \\
    Opportunity & 13.60 & 65.80 & 4.75 & 15.85 & \cellcolor{rankthird}2.95 & 24.00 & \cellcolor{rankbest}6.60 & \cellcolor{rankbest}5.80 & \cellcolor{rankthird}15.30 & \cellcolor{ranksecond}55.85 \\
    Risk & 9.30 & 73.20 & 2.65 & 14.85 & \cellcolor{ranksecond}2.85 & \cellcolor{rankthird}21.60 & \cellcolor{ranksecond}7.80 & \cellcolor{rankthird}7.30 & \cellcolor{rankbest}13.65 & \cellcolor{rankbest}56.00 \\
    \specialrule{0.4pt}{2pt}{0pt}
    \rowcolor{contextband}\multicolumn{11}{c}{\raisebox{-2pt}{\normalsize\bfseries\itshape Domains}} \\[3pt]
    \specialrule{0.4pt}{0pt}{2pt}
    Community & 10.13 & 71.00 & 2.63 & 16.25 & 3.13 & \cellcolor{rankthird}20.88 & \cellcolor{rankthird}6.63 & 7.50 & 15.00 & 53.88 \\
    Education & 15.00 & 65.50 & 2.50 & 17.00 & \cellcolor{ranksecond}2.38 & 22.13 & 9.00 & \cellcolor{ranksecond}5.88 & 15.38 & \cellcolor{ranksecond}58.50 \\
    Family & 14.38 & 60.25 & 4.25 & 21.13 & \cellcolor{rankthird}2.50 & 23.50 & 7.25 & 6.88 & 14.63 & 55.00 \\
    Health & 10.00 & 77.38 & 7.00 & 5.63 & 3.13 & 21.38 & 9.75 & 9.13 & \cellcolor{rankbest}11.88 & 53.25 \\
    Household & 8.75 & 68.25 & 3.25 & 19.75 & \cellcolor{rankbest}2.25 & \cellcolor{rankbest}18.75 & \cellcolor{rankbest}6.00 & 7.88 & \cellcolor{rankthird}13.38 & \cellcolor{rankbest}58.88 \\
    Finance & 9.63 & 71.88 & 8.50 & 10.00 & 3.38 & \cellcolor{ranksecond}19.38 & 8.25 & 8.38 & \cellcolor{ranksecond}12.50 & 55.25 \\
    Recreation & 7.63 & 80.25 & 1.50 & 10.63 & \cellcolor{ranksecond}2.38 & 22.00 & \cellcolor{ranksecond}6.50 & 8.75 & 14.25 & \cellcolor{rankthird}58.25 \\
    Social & 10.88 & 63.00 & 3.75 & 22.38 & 3.13 & 26.75 & 7.38 & 6.38 & 18.63 & 50.88 \\
    Travel & 10.75 & 71.13 & 1.38 & 16.75 & 2.75 & 26.25 & 7.25 & \cellcolor{rankbest}5.38 & 15.75 & 53.25 \\
    Work & 8.25 & 63.38 & 7.00 & 21.38 & \cellcolor{rankbest}2.25 & \cellcolor{ranksecond}19.38 & 7.38 & \cellcolor{rankthird}6.13 & 15.75 & 56.00 \\
    \bottomrule
  \end{tabular}
\end{table*}

\subsubsection{Overview}
\label{sec:response_collection}

The 8 scenarios within a personal environment share its initial state $S_0$ and have separate event sequences.
For checkpoint $t=1,\ldots,T$, the runner independently resets the environment to $S_0$ and replays only that scenario's observed prefix $(e_1,\ldots,e_t)$.
Earlier agent interventions do not affect other checkpoints or scenarios.
The agent receives protagonist context and the event prefix, using the newest timestamp as simulation time.

For human annotation, we sample 2 distinct checkpoints per model in each scenario, yielding 800 responses per model, with 200 per need and 80 per domain.
Sampling balances event positions without using human verdicts.
Details appear in Appendix~\ref{sec:sampling}.

\subsubsection{Annotation Protocol}
\label{sec:annotation_procedure}

20 annotators provide 1 verdict per response, keeping each scenario's annotator fixed across models.
Human verdicts and diagnosis use the event prefix, retrieved evidence, response, and execution result.
Appendix~\ref{sec:annotation_consistency} describes annotator preparation.

\paragraph{Action levels.}
Annotators confirm or correct the command-based action level pre-filled by the interface.
\underline{\textit{L0 Observation}} observes or retrieves information without initiating an intervention.
\underline{\textit{L1 Attention}} draws the user's attention to relevant information or needs without carrying out the underlying task.
\underline{\textit{L2 Preparation}} prepares an intervention or organizes internal information for later use without creating an external commitment.
\underline{\textit{L3 Execution}} carries out an intervention that changes the user's active commitments or creates external effects.
Higher levels do not imply better assistance.

\paragraph{Human Verdict and Diagnosis.}
\label{sec:metrics}
Annotators first give an acceptability verdict on whether a response provides grounded, useful help with appropriate timing and scope, accepting L0 when restraint is warranted.
Acceptability is the proportion of accepted responses. These automatically receive Appropriate timing and intensity.
Rejected L0 responses receive the human diagnosis \underline{\textit{Inaction}} for missing warranted help, without further diagnostic labels.
For rejected L1--L3 responses, separate questions assess timing and intensity.
\underline{\textit{Premature}} and \underline{\textit{Belated}} indicate intervention before and after its useful window, respectively.
\underline{\textit{Insufficient}} indicates inadequate help, while \underline{\textit{Excessive}} indicates intervention beyond supported scope or authority.
Appropriate denotes suitable timing or proportionate intensity.
Annotators may also flag Other Issues, detailed in Appendix~\ref{sec:other_issues}.

\paragraph{Resource Consumption.}
For each annotated response, we sum reasoning tokens across retained calls and count completed retrieval rounds.

\section{Experiments}
\label{sec:experiments}

\subsection{Settings}
\label{sec:settings}

\subsubsection{Evaluated Models}
\label{sec:models}

We evaluate 5 proprietary models: Gemini3.8-Flash and Gemini3.1-Pro \citep{google2026flash,google2026pro}, Claude-Opus5.5 \citep{anthropic2026opus}, Grok4.7 \citep{xai2026grok}, and GPT5.6-Luna \citep{openai2026luna}.
The 5 open-source models are DeepSeekV4-Pro \citep{deepseek2026v4}, GLM5.3-Flash \citep{zai2026glm}, Qwen3.5-9B \citep{qwen2026qwen35}, Qwen3.8-27B \citep{qwen2026qwen38}, and GPT-OSS-20B \citep{openai2025gptoss}.

\subsubsection{Annotation Platform}
\label{sec:annotation_platform}

We deploy an online annotation platform on Streamlit Community Cloud, with a Streamlit frontend, Python backend, and Supabase PostgreSQL storage.
Its 3 workspaces support scenario audit, response evaluation, and administration.
The 8,000 responses are annotated in 2 sessions of 4,000 each.
Annotators inspect assigned items and save or revise annotations centrally.
Interface screenshots appear in Appendix~\ref{sec:annotation_interface}.

\subsection{Main Results}
\label{sec:results}

\subsubsection{Model Comparison}
\label{sec:model_comparison}

Table~\ref{tab:main_results} compares acceptability, action levels, recorded consumption, and timing and intensity errors on the same 800 responses per model.

\paragraph{Acceptability.}
Scores range from 30.38\% for GPT-OSS-20B to 72.88\% for Gemini3.8-Flash, a 42.50-point gap.
DeepSeekV4-Pro leads open-source models with 56.63\% acceptability.
The score range reflects substantial variation in acceptability under the same annotation protocol, while even the strongest model leaves over 25\% of responses unacceptable.
Proprietary models average 63.08\%, versus 47.55\% for open-source models, a 15.53-point gap.
These means also conceal uneven performance within each group.
The 3 strongest open-source models lie within 2.50 points of each other, whereas the full group spans 26.25 points.
Their lower mean thus reflects a weak tail as well as a gap at the top.
The groups overlap: DeepSeekV4-Pro exceeds both Gemini3.1-Pro and GPT5.6-Luna, so group membership provides an incomplete account of individual quality.
These scores cover completed, human-annotated responses under the recorded inference settings, excluding transport and parsing failures.

\paragraph{Action choices.}
Attention (L1) is the most frequent level for every model, accounting for 69.20\% of all responses.
L1 exceeds 50\% of responses in 9 models and ranges from 48.00\% for Claude-Opus5.5 to 87.00\% for GPT5.6-Luna.
Preparation (L2) remains rare at 4.18\%, while execution (L3) accounts for 16.09\%.
This concentration makes the timing and usefulness of L1 responses central to the aggregate result.
At the group level, L3 shares are almost identical at about 16.1\%, despite the 15.53-point acceptability gap.
The main observed policy difference lies in observation and attention: proprietary models choose L0 about 2 times as often, while open-source models choose L1 more often.
Execution frequency alone cannot explain the observed acceptability gap.

\paragraph{Resource Consumption.}
Open-source models use fewer retrieval rounds at similar mean reasoning usage.
They average 4.77k reasoning tokens and 1.21 retrieval rounds per response, compared with 5.53k and 2.55 for proprietary models.
The reasoning-token means differ by about 14\%, whereas open-source models use about 53\% fewer retrieval rounds.
Lower recorded retrieval use accompanies lower acceptability, so resource savings should be assessed alongside the quality of the resulting responses.
Within the proprietary group, second-ranked Claude-Opus5.5 uses about 23\% of the recorded reasoning tokens of the leading Gemini3.8-Flash.
Strong performance thus occurs at different levels of recorded consumption.
These profiles support comparing quality and resource use jointly under the recorded inference settings and provider accounting conventions.

\subsubsection{Need and Domain Profiles}
\label{sec:need_domain_profiles}
\label{sec:context_results}

Table~\ref{tab:context_results} summarizes action levels, errors, and acceptability by proactive need and domain.
Acceptability ranges from 54.35\% to 56.00\% across needs and from 50.88\% to 58.88\% across domains.
Finance has more L2 preparation than Travel, and Social more L3 execution than Health, with both contrasts holding in every model.
These profiles describe sampled scenarios and available action opportunities.
Model-specific and joint profiles appear in Appendix~\ref{sec:context_findings}.

\subsubsection{Error Analysis}
\label{sec:aggregate_errors}

We identify over-proactivity as Premature timing or Excessive intensity, counting each response once when either diagnosis applies.
As shown in Figure~\ref{fig:errors}, it occurs in 2,188 of 3,575 rejected responses (61.20\%).
Model-wise shares range from 44.23\% to 79.66\%, exceeding 50\% of rejections in 6 of 10 models.
Across all responses, Premature timing is almost 3 times as common as Belated timing, while Excessive intensity is about 2 times as common as Insufficient.
These 2 asymmetries indicate more frequent early or excessive intervention.
This pattern holds across all 4 proactive need types and all 10 domains, with over-proactivity accounting for more than 50\% of rejections in each category.

The 2 highest-acceptability models also have the lowest Premature and Excessive rates, associating stronger overall performance with fewer premature or excessive interventions.
Inaction alone gives a different picture: Qwen3.8-27B records 0\% Inaction but reaches only 55.13\% acceptability and chooses L0 on just 0.38\% of responses.
Because Inaction applies only to rejected L0 responses and is reported over all responses, a policy that rarely chooses observation has few opportunities to incur this error.
Action-level shares therefore provide essential context for interpreting individual error rates.
\textbf{\textit{Over-proactivity is the dominant failure mode of proactive agents.}}

\begin{figure}[t]
  \centering
  \captionsetup{skip=4pt}
  \includegraphics[width=\columnwidth]{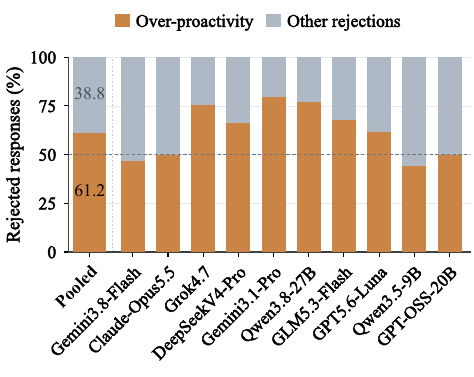}
  \caption{Over-proactivity among rejected responses.}
  \label{fig:errors}
  \vspace{-8pt}
\end{figure}

\section{Further Pattern Analysis}
\label{sec:findings}

We examine action and restraint, need-dependent strategies, evidence acquisition, and residual failures despite appropriate timing.

\subsection{Action and Restraint}
\label{sec:overall_findings}

We compare intervention choices and acceptability at shared checkpoints.

\vspace{3pt}
\begin{finding}
\raggedright
More frequent intervention can accompany lower acceptability when warranted restraint losses outweigh accepted action gains.
\end{finding}
\vspace{3pt}

\begin{figure}[t]
  \centering
  \captionsetup{skip=4pt}
  \includegraphics[width=\columnwidth]{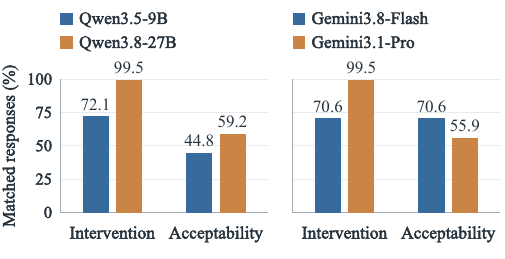}
  \caption{Paired intervention rates and acceptability.}
  \label{fig:matched_comparisons}
  \vspace{-8pt}
\end{figure}

Figure~\ref{fig:matched_comparisons} compares Qwen3.5-9B to Qwen3.8-27B at 201 shared checkpoints and Gemini3.8-Flash to Gemini3.1-Pro at 623.
Pairs match instance, scenario, event, and annotator without filtering verdicts.
Intervention frequency, the L1--L3 share, rises by 27.36 and 28.89 points, respectively, yet acceptability rises by 14.43 points for Qwen and falls by 14.77 for Gemini.
The Gemini loss combines a 5.94-point gain from accepted interventions with a 20.71-point decline from accepted L0.
Both changes use all matched responses as their denominator and sum to the acceptability change, so the intervention gain reflects a frequency change, not conditional success.

The error changes clarify the different outcomes: Qwen reduces Inaction-or-Insufficient by 17.91 points while increasing Premature-or-Excessive by 5.97, whereas Gemini increases the latter by 21.51.
Matching fixes the observed prefix and annotator, but configurations and retrieved evidence can differ, with Qwen's version and parameter count changing together.
The comparisons therefore do not isolate the causal effect of intervening more.
Restraint can also follow substantial inspection: Flash's 182 accepted L0 responses average 16.74 commands, versus 14.65 for 401 accepted interventions.
These groups cover different checkpoints, but show that L0 records final commitment rather than inspection effort.

\subsection{Needs and Strategies}
\label{sec:need_findings}

Similar acceptability across need types can mask different balances of accepted intervention and warranted restraint. Figure~\ref{fig:need_strategy} contrasts Opportunity and Dependency. Despite similar acceptability (55.85\% vs.\ 55.05\%), models are more likely to refrain from intervention for Opportunity: L0 accounts for 13.60\% of Opportunity responses, compared with 7.20\% for Dependency. Accordingly, accepted L0 contributes 10.65 points to Opportunity acceptability and 5.25 points to Dependency acceptability, while accepted interventions contribute 45.20 and 49.80 points, respectively.

All 10 models choose L0 more often for Opportunity than Dependency, although the size of this difference varies substantially. Gemini3.8-Flash chooses L0 for 39.00\% of Opportunity responses and 17.00\% of Dependency responses, whereas Qwen3.8-27B chooses L0 for 0.50\% and 0.00\%, respectively. These differences do not imply fixed action requirements for each need type: need labels characterize entire scenarios, while the appropriate action depends on the checkpoint and its available evidence. Appendix~\ref{sec:context_findings} reports model-specific acceptability profiles.

\begin{figure}[t]
  \centering
  \captionsetup{skip=4pt}
  \includegraphics[width=\columnwidth]{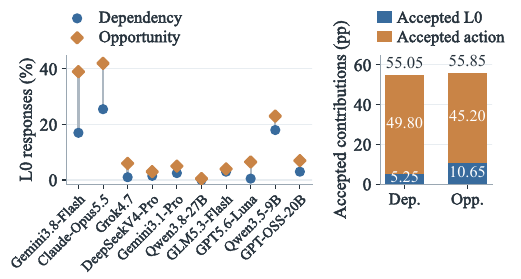}
  \caption{L0 choices by model on the left and accepted action and restraint contributions on the right.}
  \label{fig:need_strategy}
  \vspace{-8pt}
\end{figure}

\subsection{Evidence Acquisition}
\label{sec:evidence_findings}

We relate retrieval effort and local content access to human verdicts and diagnosis.

\vspace{3pt}
\begin{finding}
\raggedright
Evidence acquisition is associated with more adequate proactive assistance and higher acceptability in human verdicts.
\end{finding}
\vspace{3pt}

Figure~\ref{fig:evidence_quality} shows a positive association between mean retrieval rounds and acceptability across 10 models ($r=0.885$, $\rho=0.830$).
To examine help at fixed commitment, we compare content access among 5,536 L1 responses.
Access means obtaining non-empty database records or file text, while the no-content group includes no inspection, filenames or schemas alone, and empty results.
Acceptability is 59.0\% with content versus 40.8\% without it, with Insufficient rates of 5.2\% and 15.5\%, respectively.
Content access is thus associated with more adequate help within L1.

\begin{figure}[t]
  \centering
  \captionsetup{skip=2pt}
  \includegraphics[width=\columnwidth]{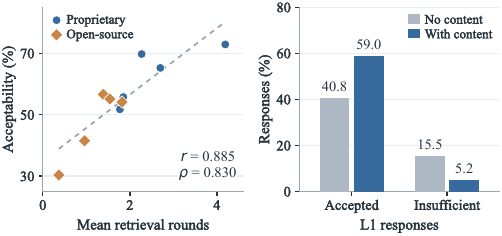}
  \vspace{-8pt}
  \caption{Mean retrieval rounds and acceptability across models (left), and L1 response proportions with and without local content access (right).}
  \label{fig:evidence_quality}
  \vspace{-8pt}
\end{figure}

After excluding singleton checkpoints, linear probability models with checkpoint fixed effects and model indicators use 5,232 L1 responses at 1,730 checkpoints.
Content access corresponds to $+9.25$ points in acceptability and $-5.86$ in Insufficient.
These associations persist when restricting the comparison to responses that inspect the environment, so the contrast is not confined to responses with versus without inspection.

\subsection{Timing and Scope}
\label{sec:error_findings}

We examine intensity failures within rejected interventions with Appropriate timing.

\vspace{3pt}
\begin{finding}
\raggedright
Among well-timed rejections, reminders tend to provide insufficient help, while executions tend to exceed acceptable scope.
\end{finding}
\vspace{3pt}

Figure~\ref{fig:timely_intensity} examines 991 rejected interventions with Appropriate timing.
Among them, 329 of 623 L1 responses (52.81\%) are Insufficient and 238 of 312 L3 responses (76.28\%) are Excessive.
The opposite intensity patterns hold within every model, although intensity does not explain all failures: 282 timely L1 rejections retain Appropriate intensity.
At the same checkpoints, 229 of the Insufficient L1 responses (69.60\%) have an accepted L1 counterpart from another model, and 178 Excessive L3 responses (74.79\%) have an accepted L1 or L2 counterpart.
Both counterpart patterns cover most corresponding rejected responses within each need.
These response-level shares depend on sampled candidates and do not measure repair success.

\begin{figure}[t]
  \centering
  \captionsetup{skip=0pt}
  \includegraphics[width=\columnwidth]{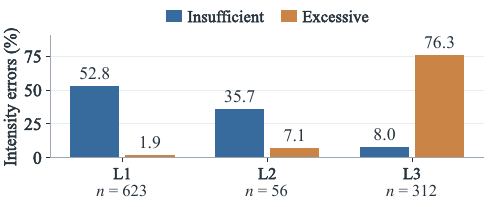}
  \caption{Intensity errors among appropriately timed but rejected interventions.}
  \label{fig:timely_intensity}
  \vspace{-12pt}
\end{figure}

Paired examples clarify the role of content and permission.
At a shared household checkpoint, Gemini3.8-Flash's accepted L1 supplies the no-pry instruction and a concrete dispatch route.
GPT-OSS-20B's Insufficient reminder instead suggests freeing the panel and supplies no usable service contact.
At a family checkpoint, the stored authorization permits a message containing only ``kitchen sneakers.''
Qwen3.8-27B sends that phrase and is accepted, while Claude-Opus5.5 adds a name, location, and Saturday-morning request and receives an Excessive diagnosis.
Both outputs are L3, and both models retrieve the wording restriction.
The same action level can thus receive different verdicts: useful detail can improve L1, while respecting narrow authorization can make L3 acceptable.

\section{Conclusion}
\label{sec:conclusion}

HALO evaluates proactive assistance through executable environments, human verdicts, and human diagnosis.
HALO-8K contains 8,000 human-annotated responses from 10 models across 400 scenarios and 2,160 checkpoints.
Over-proactivity dominates the observed rejections.
Further analyses distinguish useful action from warranted restraint, relate evidence acquisition to acceptability, and identify contrasting intensity errors among well-timed rejections.
These patterns expose calibration problems obscured by aggregate acceptability.
HALO-Arena provides operational applications and separate structured and unstructured evidence stores.
Repeated retrieval lets an agent inspect relevant evidence before deciding whether to notify, prepare, execute, or remain silent.

Acceptability peaks at 72.88\% for proprietary and 56.63\% for open-source models.
HALO records retrieved evidence and executed commands, enabling comparison of human verdicts and diagnosis for responses produced at identical checkpoints.
The practical targets differ: timely reminders need enough grounded detail to be useful, while execution must respect the scope of available permission.
These results motivate better evidence gathering and response calibration. More frequent action or higher commitment alone does not establish better assistance.
Systems should be compared jointly on acceptable action, warranted restraint, residual intensity errors, and inspection effort.

\endgroup
\clearpage
\section*{Limitations}

\paragraph{Real-world validity.}
HALO uses constructed scenarios and simulated apps to study proactive decisions under controlled conditions.
Its designed need distribution does not estimate how often those needs or failure modes occur in everyday use.
Checkpoint evaluation also omits users' responses and the cumulative effects of repeated interventions, so performance may not transfer directly to sustained deployment.

\paragraph{Human personalization.}
Human verdicts depend on preferences for initiative, useful detail, and commitment.
Shared criteria ground judgments in personal context and authority, but borderline responses can be useful to some users and intrusive to others.
The annotators' broadly similar backgrounds further limit the preferences represented, so aggregate acceptability should not be treated as a universal user preference.

\paragraph{Evaluation granularity.}
Action levels group interventions by commitment, but actions at the same level can differ in reversibility and user burden.
Human verdicts and diagnostic labels identify acceptable help and common failures without ranking acceptable alternatives by usefulness.
Consequently, aggregate acceptability does not capture the full tradeoff between benefit, burden, and intervention risk.

\begingroup
\urlstyle{same}
\def\UrlBreaks{\do\/\do\-\do\.\do\_}
\def\UrlBigBreaks{}
\let\url\nolinkurl
\bibliography{references}

@misc{zhou2023webarena,
  title = {{WebArena}: A Realistic Web Environment for Building Autonomous Agents},
  author = {Zhou, Shuyan and Xu, Frank F. and Zhu, Hao and Zhou, Xuhui and Lo, Robert and Sridhar, Abishek and Cheng, Xianyi and Ou, Tianyue and Bisk, Yonatan and Fried, Daniel and Alon, Uri and Neubig, Graham},
  year = {2023},
  note = {arXiv~preprint~2307.13854},
  url = {https://arxiv.org/abs/2307.13854}
}

@misc{yao2024taubench,
  title = {{$\tau$-bench}: A Benchmark for Tool-Agent-User Interaction in Real-World Domains},
  author = {Yao, Shunyu and Shinn, Noah and Razavi, Pedram and Narasimhan, Karthik},
  year = {2024},
  note = {arXiv~preprint~2406.12045},
  url = {https://arxiv.org/abs/2406.12045}
}

@misc{lu2024proactiveagent,
  title = {Proactive Agent: Shifting {LLM} Agents from Reactive Responses to Active Assistance},
  author = {Lu, Yaxi and Yang, Shenzhi and Qian, Cheng and Chen, Guirong and Luo, Qinyu and Wu, Yesai and Wang, Huadong and Cong, Xin and Zhang, Zhong and Lin, Yankai and Liu, Weiwen and Wang, Yasheng and Liu, Zhiyuan and Liu, Fangming and Sun, Maosong},
  year = {2024},
  note = {arXiv~preprint~2410.12361},
  url = {https://arxiv.org/abs/2410.12361}
}

@misc{pasternak2025probe,
  title = {Beyond Reactivity: Measuring Proactive Problem Solving in {LLM} Agents},
  author = {Pasternak, Gil and Rajagopal, Dheeraj and White, Julia and Atreja, Dhruv and Thomas, Matthew and Hurn-Maloney, George and Lewis, Ash},
  year = {2025},
  note = {arXiv~preprint~2510.19771},
  url = {https://arxiv.org/abs/2510.19771}
}

@misc{tang2026proagentbench,
  title = {{ProAgentBench}: Evaluating {LLM} Agents for Proactive Assistance with Real-World Data},
  author = {Tang, Yuanbo and Tang, Huaze and Cao, Tingyu and Nguyen, Lam and Zhang, Anping and Cao, Xinwen and Liu, Chunkang and Ding, Wenbo and Li, Yang},
  year = {2026},
  note = {arXiv~preprint~2602.04482},
  url = {https://arxiv.org/abs/2602.04482}
}

@inproceedings{ding2026proactor,
  title = {{ProActor}: Timing-Aware Reinforcement Learning for Proactive Task Scheduling Agents},
  author = {Ding, Lei and He, Bin and Wang, Chenguang and Liu, Yang},
  booktitle = {Proceedings of the 64th Annual Meeting of the Association for Computational Linguistics (Volume 1: Long Papers)},
  year = {2026},
  pages = {18257--18303},
  publisher = {\mbox{Association for Computational Linguistics}},
  doi = {10.18653/v1/2026.acl-long.832},
  url = {https://aclanthology.org/2026.acl-long.832/}
}

@misc{oh2026proactivitygym,
  title = {Foundations of Proactive Agents: Principles, Technical Layers, and {Proactivity-Gym}},
  author = {Oh, Jio and Do, Seunghyun and Lee, Young-Jun and Whang, Steven Euijong and Kang, Dongyeop},
  year = {2026},
  note = {arXiv~preprint~2609.37267},
  url = {https://arxiv.org/abs/2609.37267}
}

@misc{liu2026agentabstain,
  title = {{AgentAbstain}: Do {LLM} Agents Know When Not to Act?},
  author = {Liu, Xun and Zhang, Yi Evie and Kasprova, Vira and Rabbani, Parisa and Zahraei, Pardis Sadat and Zhang, Tianyu and Ebrahimpour-Boroojeny, Ali and Chandrasekaran, Varun},
  year = {2026},
  note = {arXiv~preprint~2607.10059},
  url = {https://arxiv.org/abs/2607.10059}
}

@misc{li2026proevent,
  title = {{ProEvent}: An Event-centric Benchmark for Proactive Agents},
  author = {Li, Guanzhen and Pan, Liangming and Wang, Leye},
  year = {2026},
  note = {arXiv~preprint~2607.17701},
  url = {https://arxiv.org/abs/2607.17701}
}

@misc{nathani2026pare,
  title = {Proactive Agent Research Environment: Simulating Active Users to Evaluate Proactive Assistants},
  author = {Nathani, Deepak and Zhang, Cheng and Huan, Chang and Shan, Jiaming and Yang, Yinfei and Patel, Alkesh and Gan, Zhe and Wang, William Yang and Saxon, Michael and Wang, Xin Eric},
  year = {2026},
  note = {arXiv~preprint~2604.00842},
  url = {https://arxiv.org/abs/2604.00842}
}

@misc{google2026flash,
  author = {{Google DeepMind}},
  title = {{Gemini 3.8 Flash}~Model~Card},
  year = {2026},
  url = {https://deepmind.google/models/model-cards/gemini-3-8-flash/}
}

@misc{google2026pro,
  author = {{Google DeepMind}},
  title = {{Gemini 3.1 Pro} Model Card},
  year = {2026},
  url = {https://deepmind.google/models/model-cards/gemini-3-1-pro/}
}

@misc{anthropic2026opus,
  author = {{Anthropic}},
  title = {{Claude Opus 5.5}},
  year = {2026},
  url = {https://www.anthropic.com/claude-opus-5-5}
}

@misc{xai2026grok,
  author = {{xAI}},
  title = {{Grok 4.7}: Model Documentation},
  year = {2026},
  url = {https://docs.x.ai/developers/models/grok-4.7}
}

@misc{deepseek2026v4,
  author = {{DeepSeek}},
  title = {{DeepSeek-V4} Preview},
  year = {2026},
  url = {https://deepseek.com/en/news/v4-preview/}
}

@misc{zai2026glm,
  author = {{Z.ai}},
  title = {{GLM-5.3-Flash} Model Card},
  year = {2026},
  url = {https://huggingface.co/zai-org/GLM-5.3-Flash}
}

@misc{qwen2026qwen35,
  author = {{Qwen Team}},
  title = {{Qwen3.5-9B} Model Card},
  year = {2026},
  url = {https://huggingface.co/Qwen/Qwen3.5-9B}
}

@misc{qwen2026qwen38,
  author = {{Qwen Team}},
  title = {{Qwen3.8-27B} Model Card},
  year = {2026},
  url = {https://huggingface.co/Qwen/Qwen3.8-27B}
}

@misc{openai2026luna,
  author = {{OpenAI}},
  title = {{GPT-5.6 Luna}: Model Documentation},
  year = {2026},
  url = {https://developers.openai.com/api/docs/models/gpt-5.6-luna}
}

@misc{openai2025gptoss,
  author = {{OpenAI}},
  title = {{gpt-oss-20b}: Model Documentation},
  year = {2025},
  url = {https://developers.openai.com/api/docs/models/gpt-oss-20b}
}

@misc{zhang2026pibench,
  title = {{$\pi$-Bench}: Evaluating Proactive Personal Assistant Agents in Long-Horizon Workflows},
  author = {Zhang, Haoran and Xu, Luxin and Wang, Zhilin and Gui, Runquan and Zhang, Shunkai and Lei, Haodi and He, Zihao and He, Bingsu and Qin, Chicheng and Zhu, Tong and Qu, Xiaoye and Yang, Yang and Cheng, Yu and Li, Yafu},
  year = {2026},
  note = {arXiv~preprint~2605.14678},
  url = {https://arxiv.org/abs/2605.14678}
}

@inproceedings{yang2025contextagent,
  title = {{ContextAgent}: Context-Aware Proactive {LLM} Agents with Open-World Sensory Perceptions},
  author = {Yang, Bufang and Xu, Lilin and Zeng, Liekang and Liu, Kaiwei and Jiang, Siyang and Lu, Wenrui and Chen, Hongkai and Jiang, Xiaofan and Xing, Guoliang and Yan, Zhenyu},
  booktitle = {Advances in Neural Information Processing Systems},
  volume = {38},
  year = {2025},
  doi = {10.52202/085713-5582},
  url = {https://doi.org/10.52202/085713-5582}
}

@inproceedings{yang2026fingertip,
  title = {{FingerTip 20K}: A Benchmark for Proactive and Personalized Mobile {LLM} Agents},
  author = {Yang, Qinglong and Li, Haoming and Zhao, Haotian and Yan, Xiaokai and Ding, Jingtao and Xu, Fengli and Li, Yong},
  booktitle = {The Fourteenth International Conference on Learning Representations},
  year = {2026},
  url = {https://openreview.net/forum?id=n3iFV0gLMc}
}
\endgroup
\clearpage
\appendix
\raggedbottom
\renewcommand{\topfraction}{0.9}
\renewcommand{\bottomfraction}{0.8}
\renewcommand{\dbltopfraction}{0.9}
\renewcommand{\textfraction}{0.05}
\setcounter{topnumber}{4}
\setcounter{bottomnumber}{4}
\setcounter{totalnumber}{8}
\makeatletter
\setlength{\@fptop}{0pt}
\setlength{\@fpsep}{12pt}
\setlength{\@fpbot}{0pt plus 1fil}
\setlength{\@dblfptop}{0pt}
\setlength{\@dblfpsep}{12pt}
\setlength{\@dblfpbot}{0pt plus 1fil}
\makeatother
\setcounter{dbltopnumber}{2}
\section{Benchmark Comparison}
\label{sec:benchmark_comparison}

Table~\ref{tab:benchmark_comparison} contrasts selected capabilities.
Func. Apps denotes state-changing applications, DB denotes SQL queries, and Docs denotes document retrieval.
PROBE's retrieval marks cover agent baselines, while its action score measures selection and parameters.
Human Verdicts include acceptance annotation, judge-validation decisions, and user preferences.
Human Diagnosis covers per-response timing, intensity, and other issue labels for rejections.
Consumption includes reported token usage, tool actions, interaction rounds, latency, or cost, including PARE-Bench's read actions and $\pi$-Bench's interaction turns.

\begin{table*}[htb]
  \centering
  \small
  \definecolor{benchmarkgreen}{HTML}{2E7D32}
  \newcommand{\benchmarkcheck}{\textcolor{benchmarkgreen}{$\checkmark$}}
  \caption{Selected benchmark capabilities. Checks denote explicit coverage. Dashes denote features not reported in the evaluated protocol.}
  \label{tab:benchmark_comparison}
  \setlength{\tabcolsep}{3pt}
  \renewcommand{\arraystretch}{1.15}
  \begin{tabular}{@{}l*{6}{c}@{}}
    \toprule
    Benchmark & \multicolumn{3}{c}{Environment} & \multicolumn{3}{c}{Evaluation} \\
    \cmidrule(lr){2-4}\cmidrule(lr){5-7}
    & Func. Apps & DB & Docs & Human Verdicts & Human Diagnosis & Consumption \\
    \midrule
    PROBE \citep{pasternak2025probe} & -- & \benchmarkcheck & \benchmarkcheck & \benchmarkcheck & -- & -- \\
    ProAgentBench \citep{tang2026proagentbench} & -- & -- & -- & -- & -- & \benchmarkcheck \\
    PARE-Bench \citep{nathani2026pare} & \benchmarkcheck & -- & \benchmarkcheck & -- & -- & \benchmarkcheck \\
    $\pi$-Bench \citep{zhang2026pibench} & \benchmarkcheck & -- & \benchmarkcheck & -- & -- & \benchmarkcheck \\
    Proactivity-Gym \citep{oh2026proactivitygym} & \benchmarkcheck & -- & \benchmarkcheck & \benchmarkcheck & -- & -- \\
    \midrule
    \textbf{HALO (Ours)} & \benchmarkcheck & \benchmarkcheck & \benchmarkcheck & \benchmarkcheck & \benchmarkcheck & \benchmarkcheck \\
    \bottomrule
  \end{tabular}
  \par\vspace{-8pt}
\end{table*}

\section{Applications and Commands}
\label{sec:app_commands}

Table~\ref{tab:app_commands} lists app functions and their automatic action levels, with Assistant denoting the user-notification surface.
No action and interaction-only responses map to L0.
Of the 19 action commands, 6 map to L1, 6 to L2, and 7 to L3, providing comparable coverage across intervention levels.

\begin{table*}[htb]
  \centering
  \small
  \caption{Agent functions by app and category. Superscripts indicate automatically inferred intensity levels, provided to guide human verification and reduce verification time.}
  \label{tab:app_commands}
  \setlength{\tabcolsep}{3pt}
  \renewcommand{\arraystretch}{1.10}
  \begin{tabular}{@{}>{\raggedright\arraybackslash}p{0.16\textwidth}>{\raggedright\arraybackslash}p{0.42\textwidth}>{\raggedright\arraybackslash}p{0.39\textwidth}@{}}
    \toprule
    App & Interaction commands (29) & Action commands (19) \\
    \midrule
    Email & \texttt{get\_email}, \texttt{list\_emails}, \texttt{search\_emails}, \texttt{list\_drafts} & \texttt{flag\_email}\textsuperscript{L1}, \texttt{create\_draft}\textsuperscript{L2}, \texttt{send\_email}\textsuperscript{L3} \\
    Calendar & \texttt{get\_event}, \texttt{list\_events}, \texttt{find\_conflicts}, \texttt{get\_event\_draft}, \texttt{list\_event\_drafts}, \texttt{get\_reminder}, \texttt{list\_reminders} & \texttt{create\_reminder}\textsuperscript{L1}, \texttt{create\_event\_draft}\textsuperscript{L2}, \texttt{create\_event}\textsuperscript{L3}, \texttt{update\_event}\textsuperscript{L3}, \texttt{delete\_event}\textsuperscript{L3} \\
    Chat & \texttt{get\_conversation}, \texttt{list\_conversations}, \texttt{get\_message}, \texttt{list\_messages}, \texttt{search\_messages}, \texttt{list\_drafts} & \texttt{pin\_conversation}\textsuperscript{L1}, \texttt{create\_draft}\textsuperscript{L2}, \texttt{send\_message}\textsuperscript{L3} \\
    Contacts & \texttt{get\_contact}, \texttt{list\_contacts}, \texttt{search\_contacts} & \texttt{flag\_contact}\textsuperscript{L1}, \texttt{create\_contact}\textsuperscript{L2}, \texttt{update\_contact}\textsuperscript{L2} \\
    Wallet & \texttt{get\_wallet}, \texttt{get\_transaction}, \texttt{list\_transactions}, \texttt{list\_drafts} & \texttt{flag\_transaction}\textsuperscript{L1}, \texttt{create\_draft}\textsuperscript{L2}, \texttt{pay}\textsuperscript{L3}, \texttt{refund\_transaction}\textsuperscript{L3} \\
    Database & \texttt{get\_schema}, \texttt{execute} & -- \\
    Files & \texttt{get\_file}, \texttt{list\_files}, \texttt{search\_files} & -- \\
    Assistant & -- & \texttt{notify\_user}\textsuperscript{L1} \\
    \bottomrule
  \end{tabular}
  \par\vspace{-8pt}
\end{table*}

\section{Construction Details}
\label{sec:construction_details}

\subsection{Automated Checks}
\label{sec:automated_checks}

Candidates pass a deterministic Validator followed by an LLM Reviewer.
Each scenario permits up to 5 candidates, and only accepted candidates commit initialization facts and events.
Table~\ref{tab:construction_acceptance} covers 568 persisted, parseable candidates, excluding malformed outputs and transport retries.
Reviewer rates condition on 509 Validator passes, while overall rates use all 568 candidates.
Revised candidates are counted separately.
Counts are historical project records because the original candidate archive is no longer retained.

\begin{table}[htb]
  \centering
  \small
  \caption{Initial pipeline pass and fail rates. Reviewer rates are conditional on Validator acceptance.}
  \label{tab:construction_acceptance}
  \setlength{\tabcolsep}{2pt}
  \begin{tabular}{@{}lcccc@{}}
    \toprule
    Stage & Pass & Total & Pass (\%) & Fail (\%) \\
    \midrule
    Validator & 509 & 568 & 89.61 & 10.39 \\
    Reviewer & 400 & 509 & 78.59 & 21.41 \\
    Overall & 400 & 568 & 70.42 & 29.58 \\
    \bottomrule
  \end{tabular}
\end{table}

After all 8 scenarios of an instance pass, the Noise Creator adds settled, causally irrelevant records to shared initialization, expanding the inspection space without introducing unresolved activities.

\subsection{Human Audit}
\label{sec:human_audit_details}

Table~\ref{tab:scenario_audit} reports individual votes, review pairs, and adjudication outcomes.
Vote rates use 800 verdicts, while review-pair and adjudication rates use 400 scenarios.
Scenarios approved by both reviewers were retained.
Cases with 2 Needs Revision votes or 1 Needs Revision and 1 Reject vote entered revision directly.
Revisions followed reviewer feedback, prioritizing scenario corrections.
Shared environments were adjusted only when these corrections were insufficient, to limit effects on unrelated scenarios.

\begin{table}[htb]
  \centering
  \small
  \caption{Human audit and adjudication. Review pairs are unordered.}
  \label{tab:scenario_audit}
  \setlength{\tabcolsep}{2pt}
  \begin{tabular}{@{}lcc@{}}
    \toprule
    Record & Count & Rate (\%) \\
    \midrule
    \multicolumn{3}{@{}l@{}}{\textbf{Audit Votes}} \\
    Approve & 593 & 74.13 \\
    Needs Revision & 175 & 21.88 \\
    Reject & 32 & 4.00 \\
    \midrule
    \multicolumn{3}{@{}l@{}}{\textbf{Review Pairs}} \\
    Approve + Approve & 214 & 53.50 \\
    Approve + Needs Revision & 138 & 34.50 \\
    Approve + Reject & 27 & 6.75 \\
    Needs Revision + Needs Revision & 16 & 4.00 \\
    Needs Revision + Reject & 5 & 1.25 \\
    Reject + Reject & 0 & 0.00 \\
    \midrule
    \multicolumn{3}{@{}l@{}}{\textbf{Adjudication}} \\
    Approved by third reviewer & 75 & 18.75 \\
    Retained after adjudication & 289 & 72.25 \\
    Selected for revision & 111 & 27.75 \\
    \bottomrule
  \end{tabular}
\end{table}

\section{Checkpoint Selection}
\label{sec:sampling}

Table~\ref{tab:checkpoint_distribution} reports checkpoint counts by scenario length.
Within each 5-model batch, randomized maximum-flow sampling selects 2 distinct completed checkpoints per model and scenario, without using verdicts.
The resulting 10 responses cover all event positions with counts differing by at most 1.
Table~\ref{tab:checkpoint_selection} reports event positions and diagnostics.
Wasserstein distances use normalized position $(t-1)/(T-1)$ for event index $t$ and trajectory length $T$.

A linear probability model with checkpoint fixed effects and model indicators preserves the ranking, with a maximum mean-centered adjustment of 1.69 points.
This suggests limited selection sensitivity, without establishing equal difficulty or resolving small score gaps.

\begin{table}[htb]
  \centering
  \small
  \caption{Checkpoint distribution by scenario length.}
  \label{tab:checkpoint_distribution}
  \setlength{\tabcolsep}{4pt}
  \renewcommand{\arraystretch}{1.12}
  \begin{tabular}{@{}ccc@{}}
    \toprule
    Events & Scenarios & Checkpoints \\
    \midrule
    4 & 14 & 56 \\
    5 & 212 & 1,060 \\
    6 & 174 & 1,044 \\
    \midrule
    Total & 400 & 2,160 \\
    \bottomrule
  \end{tabular}
\end{table}

\begin{table}[htb]
  \centering
  \small
  \caption{Selected event positions and selection diagnostics. pp denotes percentage points.}
  \label{tab:checkpoint_selection}
  \setlength{\tabcolsep}{2pt}
  \renewcommand{\arraystretch}{1.12}
  \begin{tabular}{@{}l*{6}{c}@{}}
    \toprule
    Model & \multicolumn{6}{c}{Selected positions (\%)} \\
    \cmidrule(lr){2-7}
    & 1 & 2 & 3 & 4 & 5 & 6 \\
    \midrule
    Gemini3.8-Flash & 15.75 & 19.25 & 19.75 & 21.63 & 17.25 & 6.38 \\
    Claude-Opus5.5 & 17.63 & 19.50 & 18.88 & 19.50 & 17.25 & 7.25 \\
    Grok4.7 & 17.75 & 16.63 & 19.25 & 20.13 & 19.00 & 7.25 \\
    Gemini3.1-Pro & 15.75 & 20.00 & 21.38 & 19.00 & 17.63 & 6.25 \\
    GPT5.6-Luna & 19.38 & 19.38 & 16.88 & 18.75 & 17.25 & 8.38 \\
    DeepSeekV4-Pro & 18.00 & 18.50 & 18.25 & 19.88 & 17.88 & 7.50 \\
    Qwen3.8-27B & 18.25 & 16.75 & 20.88 & 19.75 & 17.25 & 7.13 \\
    GLM5.3-Flash & 17.25 & 20.38 & 19.88 & 17.88 & 18.38 & 6.25 \\
    Qwen3.5-9B & 17.75 & 20.13 & 16.75 & 20.38 & 16.88 & 8.13 \\
    GPT-OSS-20B & 17.88 & 18.88 & 17.88 & 18.63 & 19.88 & 6.88 \\
    \midrule
    All models & 17.54 & 18.94 & 18.98 & 19.55 & 17.86 & 7.14 \\
    \bottomrule
  \end{tabular}
  \par\vspace{6pt}
  \begin{tabular}{@{}lc@{}}
    \toprule
    Selection diagnostic & Value \\
    \midrule
    Cram\'er's $V$ & 0.029 \\
    Max. pairwise Wasserstein distance & 0.0288 \\
    Max. centered score adjustment & 1.69 pp \\
    Max. pairwise score-gap change & 3.27 pp \\
    Adjusted model ranking & Unchanged \\
    \bottomrule
  \end{tabular}
\end{table}

\section{Annotation Consistency}
\label{sec:annotation_consistency}


\label{sec:annotation_preparation}

2077AI oversaw annotator recruitment, screening, management, and ongoing follow-up to support consistent application of the annotation guidelines.
A shared briefing and dry run familiarized 20 annotators with the rubric and interface.
Annotators had broadly similar demographic backgrounds.




\section{Pairwise Comparison}
\label{sec:context_findings}

\begin{figure}[htb]
  \centering
  \captionsetup{skip=4pt}
  \includegraphics[width=\columnwidth]{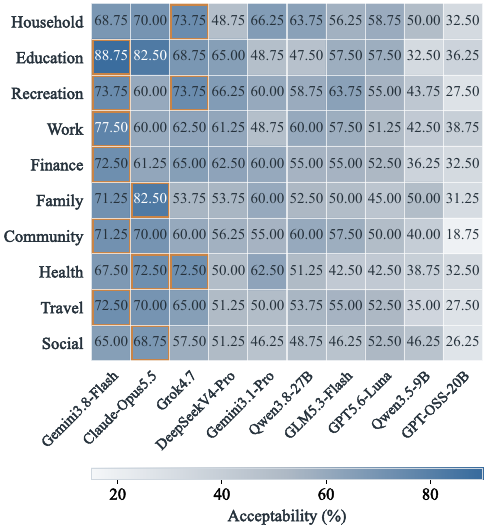}
  \caption{Acceptability by model and domain. Outlines mark domain leaders.}
  \label{fig:model_domain}
\end{figure}

\begin{figure}[htb]
  \centering
  \captionsetup{skip=4pt}
  \includegraphics[width=\columnwidth]{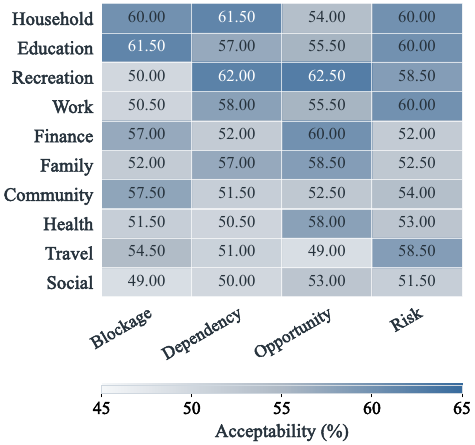}
  \caption{Acceptability by need and domain.}
  \label{fig:need_domain}
\end{figure}

Table~\ref{tab:need_acceptability} and Figures~\ref{fig:model_domain} and~\ref{fig:need_domain} report model--need, model--domain, and need--domain acceptability, respectively.
Model--domain outlines mark the best unrounded scores, including ties.
Need orderings vary by domain: Opportunity exceeds Risk in Finance, with the reverse in Travel.

\begin{table}[htb]
  \centering
  \small
  \caption{Acceptability (\%) by model and proactive need type.}
  \label{tab:need_acceptability}
  \setlength{\tabcolsep}{1.5pt}
  \renewcommand{\arraystretch}{1.12}
  \begin{tabular*}{\columnwidth}{@{\extracolsep{\fill}}lcccc@{}}
    \toprule
    Model & \footnotesize Blockage & \footnotesize Dependency & \footnotesize Opportunity & \footnotesize Risk \\
    \midrule
    Gemini3.8-Flash & 73.0 & 70.0 & 73.5 & 75.0 \\
    Claude-Opus5.5 & 69.5 & 68.5 & 70.0 & 71.0 \\
    Grok4.7 & 63.5 & 65.5 & 64.0 & 68.0 \\
    DeepSeekV4-Pro & 53.5 & 54.0 & 57.5 & 61.5 \\
    Gemini3.1-Pro & 54.0 & 57.0 & 54.0 & 58.0 \\
    Qwen3.8-27B & 54.0 & 59.5 & 52.5 & 54.5 \\
    GLM5.3-Flash & 54.5 & 51.0 & 53.5 & 57.5 \\
    GPT5.6-Luna & 49.0 & 53.0 & 50.5 & 54.5 \\
    Qwen3.5-9B & 43.0 & 40.5 & 49.0 & 33.5 \\
    GPT-OSS-20B & 29.5 & 31.5 & 34.0 & 26.5 \\
    \bottomrule
  \end{tabular*}
\end{table}

\section{Other Issues}
\label{sec:other_issues}

Other Issues are recorded for 1,000 of 3,575 rejected responses (27.97\%), corresponding to 12.50\% of all 8,000 responses.
Table~\ref{tab:other_issues} reports the category distribution.
Each flagged response has 1 recorded category, which can accompany timing or intensity errors.
Others requires a written explanation.
We retain the exported categories without reclassifying free-text comments.
Unsupported Evidence and Irrelevant Action together account for 79.40\% of Other Issues flags, linking human diagnosis to grounding and relevance.
These categories supplement timing and intensity by distinguishing unsupported content from actions that do not address the current need.

\begin{table}[htb]
  \centering
  \small
  \caption{Other Issues distribution. Within and Rejected are percentages of the 1,000 responses with Other Issues and all 3,575 rejected responses, respectively.}
  \label{tab:other_issues}
  \setlength{\tabcolsep}{2pt}
  \renewcommand{\arraystretch}{1.12}
  \begin{tabular}{@{}lccc@{}}
    \toprule
    Category & Count & Within (\%) & Rejected (\%) \\
    \midrule
    Unsupported Evidence & 503 & 50.30 & 14.07 \\
    Irrelevant Action & 291 & 29.10 & 8.14 \\
    Factual Contradiction & 128 & 12.80 & 3.58 \\
    Others & 78 & 7.80 & 2.18 \\
    \bottomrule
  \end{tabular}
\end{table}

\section{Potential of HALO}
\label{sec:future_directions}

We adapt Qwen3.5-9B and Qwen3.5-4B with non-thinking LoRA to predict human verdicts and diagnostic labels from 7,600 training responses.
The 400 validation items, 1 per scenario, share scenarios with training and select the lowest-loss checkpoint.
Both models train for 1 epoch with effective batch size 16, using microbatch size and gradient accumulation of 8 and 2 for 9B, and 2 and 8 for 4B.
Original and adapted models use identical prompts and decoding.
Intensity excludes nulls, while Joint requires all 3 labels, including nulls, to match.
Changes use unrounded scores.

\begin{table}[htb]
  \centering
  \small
  \caption{Qwen3.5 judge accuracy (\%).}
  \label{tab:judge_adaptation}
  \setlength{\tabcolsep}{4pt}
  \renewcommand{\arraystretch}{1.20}
  \begin{tabular}{@{}lcccc@{}}
    \toprule
    Measure & $N$ & Original & LoRA & Change \\
    \midrule
    \multicolumn{5}{c}{Qwen3.5-9B} \\
    \midrule
    Human verdict & 400 & 48.50 & 56.00 & $+7.50$ \\
    Timing & 400 & 61.25 & 57.25 & $-4.00$ \\
    Intensity & 389 & 57.07 & 41.90 & $-15.17$ \\
    Joint & 400 & 33.75 & 22.75 & $-11.00$ \\
    \midrule
    \multicolumn{5}{c}{Qwen3.5-4B} \\
    \midrule
    Human verdict & 400 & 51.25 & 54.75 & $+3.50$ \\
    Timing & 400 & 66.50 & 60.00 & $-6.50$ \\
    Intensity & 389 & 67.10 & 27.76 & $-39.33$ \\
    Joint & 400 & 44.75 & 16.00 & $-28.75$ \\
    \bottomrule
  \end{tabular}
\end{table}

Table~\ref{tab:judge_adaptation} shows verdict accuracy gains of 7.50 and 3.50 points for 9B and 4B, alongside joint accuracy declines of 11.00 and 28.75 points.
Verdict gains can therefore mask diagnostic regressions.
These are exploratory validation comparisons, not independent test results, and do not isolate effects of class weighting or model size.


\section{Prompts}
\label{sec:prompts}

We provide prompt templates.

\subsection{Agent Prompt}
\begin{haloprompt}
You are a proactive personal Agent in a simulated environment.

At this checkpoint, decide whether to keep observing, inspect local information,
or proactively help. Judge only the observed Event prefix; checkpoints are
independent, so no earlier Agent behavior is part of the state. Use only observed
Events and inspection results. Do not infer future Events or invent records.
Intervene only when it is useful now. Choose an intervention that is effective
for the current need, supported by the available evidence, and safe. This
judgment is internal and must not be described in the JSON.

Events may reveal the need without containing every fact required for a good
response. When a possible intervention depends on a qualification, prior
record, identifier, procedure, constraint, preference, inventory fact, or safe
next step, inspect the local SQLite database and/or TXT Files before responding.
Do not assume that communications and calendar entries are the complete source
of truth, and do not invent a missing detail instead of retrieving it.

Use assistant.notify_user when a grounded message to the Instance user would
provide useful help now. Explain the relevant finding or concrete next step;
merely restating Events or giving generic advice is insufficient. The message
does not obtain a user reply or approval, or carry out the operations it
suggests. Never claim to have completed an operation that was not performed.

The time field of the newest observed Event is the current simulation time. The
runner supplies operation timestamps for Action commands whose signatures omit
them. App record timestamps and Calendar business times do not replace the
Event time. Ignore the real wall-clock date and any model training date; an
Event is not "in the past" merely because its absolute date precedes the real
date outside this episode.

Return one compact JSON object in exactly one of these forms:
{
  "decision": "inspect",
  "content": [{"app":"...","function":"...","parameter":...}]
}
or
{
  "decision": "final",
  "content": null | {"app":"...","function":"...","parameter":...}
}

Rules:
- inspect: content must be one non-empty list of private information-gathering
  commands from the Inspect commands below. The runner executes these commands
  and returns their results at the next round. You may repeat retrieval as often
  as the evidence requires; there is no fixed retrieval limit.
- final: content must be null for no action or exactly one intended command
  from the Action commands below. The command is executed once at the isolated
  checkpoint to verify its result. Never return multiple final
  commands or a standalone text response.
- Null final content applies only to this checkpoint.

Instance context:
<<INSTANCE>>

Observed Events through the current checkpoint:
<<EVENTS>>

Inspection results from this checkpoint:
<<TOOL_RESULTS>>

Agent-facing App surface:
<<TOOLS>>
\end{haloprompt}

\subsection{Construction Prompts}

\subsubsection{Instance Creator}
\begin{haloprompt}
You create base Instances for a proactive-agent benchmark.

Given a broad Domain, define one coherent person and a concrete part of that
person's life. The Instance should be narrower than the Domain while supporting
eight meaningfully different Scenarios across several recurring activities.
Create stable background state, not a proactive need.

Keep the setting understandable to a general reader and valid throughout
calendar year <<BENCHMARK_YEAR>>. Make it materially different from existing
Instances in role, responsibilities, workflows, and record design.

Return raw JSON only, without Markdown or code fences. Return one JSON object
with exactly these fields:
1. instance_id: copy the assigned ID.
2. instance_description: roughly 100 words total, written as four concise,
   newline-separated items using exactly these labels: Setting, Scope,
   Resources, and Exclusions. Cover the setting, recurring activity areas,
   available records, and Instance boundaries needed for later construction.
   Format every item as `Label: text` without bullets or Markdown.
3. personal_description: roughly 100 words total, written as four concise,
   newline-separated items using exactly these labels: Identity,
   Responsibilities, Authority, and Restrictions. Define one protagonist whose
   full name is unique in the benchmark and the stable personal context
   relevant across its Scenarios. Later Scenario data will use that name
   verbatim as user_id.
   Format every item as `Label: text` without bullets or Markdown.
4. commands: initialization commands.

Define 2-3 empty SQLite tables with at most 5 columns each, including one table
that can hold protagonist-related records; define 2-3 reusable empty TXT Files;
and initialize exactly one zero-balance Wallet. The Instance establishes only
reusable structure: use CREATE statements only, add no database rows, and put
no content in the Files. Scenario-specific facts and File content belong in
later Scenario initialization. Put all table and File definitions before the
Wallet command. Give every table and File definition a description, and keep
native App state out of the database.

Use the exact lowercase App functions and argument shapes in the App surface.
Each command has app, function, parameter, and description only when required
for a table or File definition.

Assigned Domain:
<<DOMAIN_SPEC>>

Assigned Instance ID: <<INSTANCE_ID>>

Existing Instances in this Domain:
<<EXISTING_INSTANCES>>

Protagonist names already used elsewhere:
<<USED_PROTAGONIST_NAMES>>

Retry feedback:
<<RETRY_FEEDBACK>>

App surface:
<<APP_SURFACE>>
\end{haloprompt}

\subsubsection{Scenario Creator}
\begin{haloprompt}
You create Scenarios for a proactive-agent benchmark.

Create one <<TRIGGER_TYPE>> Scenario.
Definition: <<TRIGGER_DEFINITION>>

Choose the causal design by the state of the need across the evaluated prefixes,
not by its surface stakes. If another trigger definition fits that state more
directly, redesign the Scenario rather than relabel it.

Return exactly one raw JSON object and nothing else: no Markdown, fences,
commentary, or second object. Its top level contains only scenario_description,
events, and ini_data_commands. Choose the trajectory length for this Scenario
within the benchmark's 4-to-6 Event range. events must be one JSON object with
sequential keys event1 onward; it must not be a list or be expanded into
top-level Event fields.
scenario_description is roughly 100 words total, written as four concise,
newline-separated items using exactly these labels: Objective, Emerging Need,
Decision Context, and Consequence. Cover the causal situation and intended
trajectory without retelling every Event, prescribing the expected Agent
action, or exposing the hidden design checks. Format every item as
`Label: text` without bullets or Markdown.
Keep all four items Scenario-specific. Do not restate the protagonist's
identity, standing responsibilities or authority, the general Instance scope,
or its resource inventory. Mention a stable fact only when its application to
this particular trajectory is causally necessary.
Every Event contains exactly time, app, function, and the singular field
parameter. time is the ISO 8601 time, including a timezone, when that Event
enters the observable environment. Event times must be strictly increasing.
An App record's parameter.timestamp is separate: it describes the record and
cannot be later than the Event time. Calendar start, end, reminder, and due
times remain business times rather than Event times. Every initialization
command contains exactly app, function, and parameter. The program adds
scenario_id, trigger_type, and reference_time, so do not return them. Use the
assigned reference date as the earliest Event date.

Design requirements:
- Build one causal trajectory around exactly one need of the assigned trigger
  type. Every Event must develop that same need by changing its evidence,
  feasibility, options, or appropriate response; do not introduce a second
  proactive need or a competing trigger.
- As silent design checks, ensure that at least one Event prefix would make the
  contemplated intervention premature and therefore unacceptable, and that at
  least one later prefix would make intervention unacceptable because acting
  only then is already too late to achieve its intended benefit or prevention.
  Both judgments must follow from evidence in their respective prefixes. Do not
  state timing labels, prefix verdicts, or this design reasoning anywhere in the
  output JSON; present only the Scenario's world state and observable Events.
- Events are observable changes in the world. They may supply evidence,
  urgency, options, or confirmation, but must not contain the evaluated Agent's
  action, resolve the need, or turn the intervention into a direct response to
  an explicit request. Information may move between actors or Apps only through
  prior shared state or an observable communication in the prefix.
- The best intervention must require at least one decision-relevant fact from
  the Instance SQLite database or a TXT File. That fact must distinguish
  plausible recognition, timing, or action choices rather than merely support
  a conclusion already forced by Events. Events must neither state that fact
  nor independently imply the same decision. Use ini_data_commands when the
  shared state needs additional pre-Event facts or prerequisites.
- The contemplated intervention must materially improve the protagonist's
  situation when supported by the Event state, match the trigger definition,
  and have a commitment level proportionate to the available authority,
  consent, and reversibility. Every fact and parameter needed for that
  intervention must be available in the relevant prefix; when authority or
  preference is missing,
  use a lower-commitment step that preserves the protagonist's options. Trace
  each decisive condition for timing and action to a source available in that
  prefix; if one is missing or appears only later, revise the action or timing
  design. For blockage, the same activity must be shown as started and then
  actually unable to continue.
- Use at least three distinct non-retrieval Apps in the Event sequence. Do not
  count database, files, initialization, or information-retrieval steps toward
  this minimum. Every counted App must supply necessary causal evidence or a
  state transition for the single need; if three Apps do not fit naturally,
  redesign the causal progression rather than adding a decorative channel.
- Compare the accepted summaries before choosing the causal design. Make the
  Scenario materially different in its goal, cause, key information, Event
  progression, and likely intervention; changing surface details within the
  same workflow is not a material difference. A new Trigger label or failure
  mechanism alone does not make reuse of the same underlying activity and
  protagonist goal distinct. Compare the causal spine as well: if the same
  state-transition pattern leads to the same intervention, changing its subject
  matter is still a surface variation. Across the full set, distribute activity
  areas, retrieval sources, Event channels, and useful Agent actions.

State and consistency requirements:
- ini_data_commands describe only facts that exist before event1. Preserve the
  protagonist, Wallet, accepted shared state, existing SQL schemas, and natural
  key uniqueness. Keep every dated initialization fact on or before the assigned
  reference date. Add no unrelated noise.
- Use native personal Apps for their own data. Information-retrieval calls are
  not Events. send_email and send_message are evaluated-Agent actions, not
  Scenario Events or initialization commands.
- Match every command's parameter keys to its exact App signature. An Event's
  time does not add a timestamp argument to a function that does not define one.
- The eight Scenarios share accumulated initialization data, while their Event
  sequences are independent branches and are never each other's history.
- Correct every retry issue with a genuinely revised candidate rather than a
  renamed version of the same causal design.

Before returning, inspect the assembled JSON for contract integrity across its
keys, collection shapes, counts, and member schemas, then check the causal
requirements above. Contrast the assigned trigger with its closest alternative
across the Event sequence; if the distinction depends on wording rather than
the state of the need, redesign the Scenario.

Assigned Scenario ID: <<SCENARIO_ID>>
Assigned reference date: <<REFERENCE_DATE>>
Protagonist: <<PERSONAL_DESCRIPTION>>

Current shared Instance state:
<<INSTANCE_CONTEXT>>

Accepted Scenarios with the same trigger type:
<<SAME_TRIGGER_SCENARIOS>>

Accepted Scenarios with other trigger types:
<<OTHER_SCENARIOS>>

Retry feedback:
<<RETRY_FEEDBACK>>

App surface:
<<APP_SURFACE>>
\end{haloprompt}

\subsubsection{Reviewer}
\begin{haloprompt}
You perform semantic review for a proactive-agent benchmark.

The deterministic Validator has already checked JSON structure and isolated App
execution. The candidate is not committed: its initialization commands would
join the shared initial state, while every Scenario's Events remain an
independent branch. The candidate intentionally does not reveal an expected
Agent answer or timing labels. Infer the concrete, most useful intervention
that the observable state and available tools support; it must not already
occur in Events or initialization. Reject when no such intervention is
reasonably inferable or feasible, not merely because the JSON does not state
the answer explicitly.

Return raw JSON only, without Markdown or code fences. Return one JSON object
with exactly:
{"valid": boolean, "issues": [concise, concrete strings]}

Judge only material benchmark quality:
1. Trigger fidelity: the evidence matches the expected trigger definition.
2. Timing and observability: the sequence supports at least one prefix where
   action would be unacceptably premature and at least one later prefix where
   acting only then would be unacceptably late. Treat each Event's time as the
   current time at that prefix, distinct from App record and business times.
   Judge each boundary only from evidence available in that prefix.
3. Retrieval dependency: a necessary database or TXT fact is available, not
   duplicated by Events, and materially changes recognition, choice, or action.
4. Intervention value: the contemplated help is useful, feasible when supported
   by the Event state, proportionate to the available authority, consent, and
   reversibility, and not already supplied by the normal process or by an
   explicit request containing the answer.
5. Causal and shared-state integrity: prerequisites exist, initialization is
   genuinely pre-Event, external developments have causes, and all shared facts
   can coexist. Do not treat Events from independent Scenarios as shared history.
6. Set quality: the candidate adds meaningful diversity in activity, cause,
   information source, Event progression, and likely intervention without using
   irrelevant Apps or merely renaming an accepted design. A different Trigger
   label or failure mechanism does not by itself justify repeating the same
   underlying activity and protagonist goal; the causal spine and likely
   intervention must also be materially different from the nearest example.

Trigger definitions: dependency=future prerequisite missing; blockage=an active
attempt has actually stopped; risk=plausible future loss has not happened yet;
opportunity=optional benefit with no failure or loss from inaction.

Do not reject harmless additions to sparse base state, stylistic preferences,
or set-level quotas that cannot yet be satisfied during an intermediate round.
A candidate is valid only when no material issue remains.

Expected candidate trigger: <<EXPECTED_TRIGGER>>
Domain: <<DOMAIN_SPEC>>
Instance: <<INSTANCE_PROFILE>>

Accepted Scenarios:
<<ACCEPTED_SCENARIOS>>

Candidate Scenario, including initialization:
<<CANDIDATE_SCENARIO>>

Shared state before the candidate:
<<BASE_STATE>>
\end{haloprompt}

\subsubsection{Noise Creator}
\begin{haloprompt}
You create background Noise for a proactive-agent benchmark.

All eight Scenarios are complete. Return raw JSON only, without Markdown or
code fences. Return one JSON object with exactly:
{"noise_commands": [...]}

Create 12-20 valid app/function/parameter commands that add ordinary,
Domain-coherent background state across several Apps. Use existing database
schemas, exact App signatures, unique natural keys, affordable completed Wallet
activity, and valid Messaging conversations.

Noise must remain causally irrelevant to every accepted Scenario. It must be
settled or inert at the initial state: it cannot introduce evidence, urgency,
an unresolved need, a future obligation, a pending outcome, or an evaluated
Agent action. Preserve all existing objects, identity, Wallet configuration,
and Scenario facts.

Protagonist: <<PERSONAL_DESCRIPTION>>

Final shared state before Noise:
<<INSTANCE_CONTEXT>>

Accepted Scenarios:
<<ACCEPTED_SCENARIOS>>

Retry feedback:
<<RETRY_FEEDBACK>>

App surface:
<<APP_SURFACE>>
\end{haloprompt}

\section{Annotation Interface}
\label{sec:annotation_interface}

\begin{figure*}[htb]
  \centering
  \captionsetup{skip=4pt}
  \includegraphics[width=0.77\textwidth]{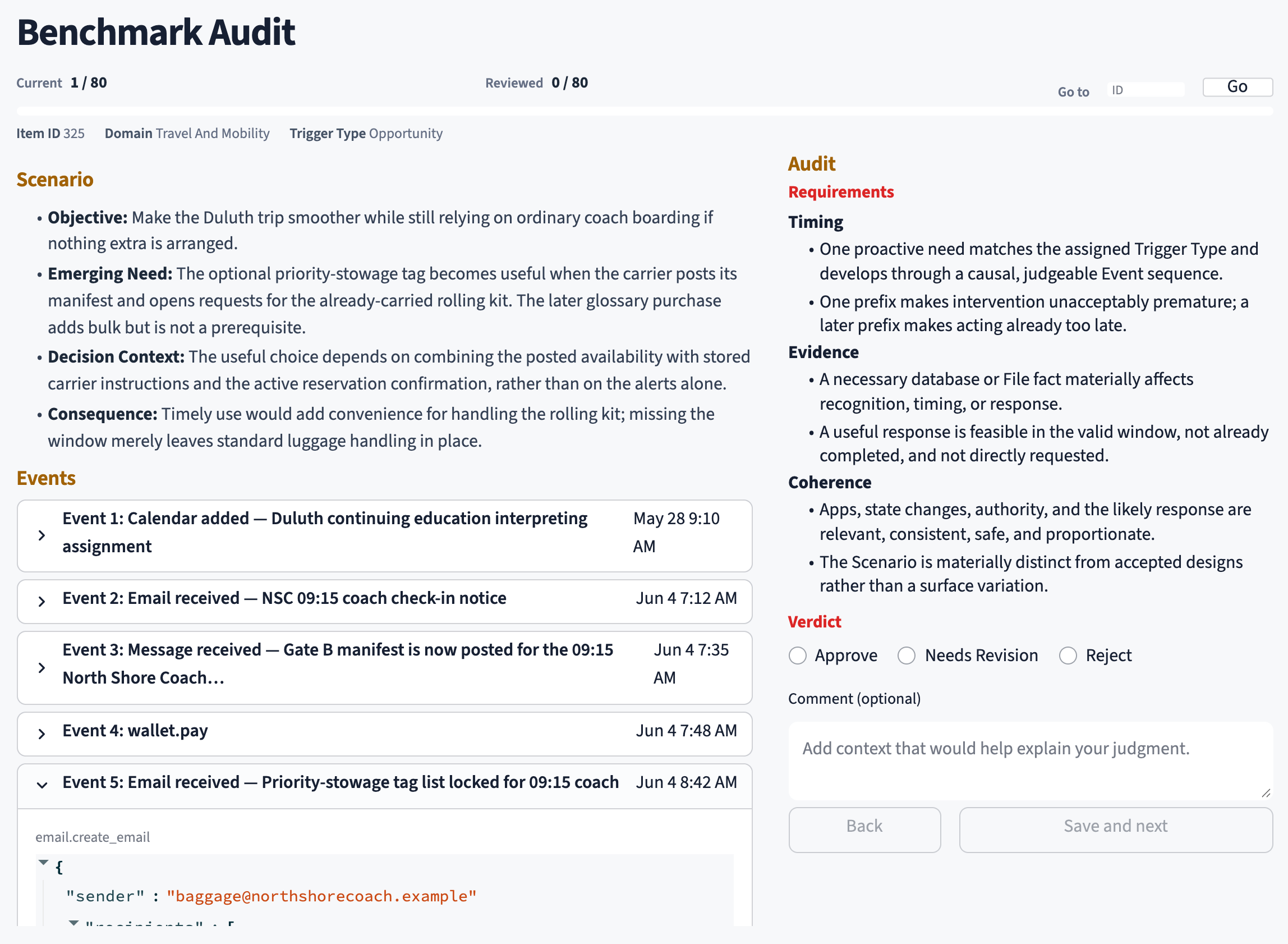}\par
  \vspace{6pt}
  \includegraphics[width=0.77\textwidth]{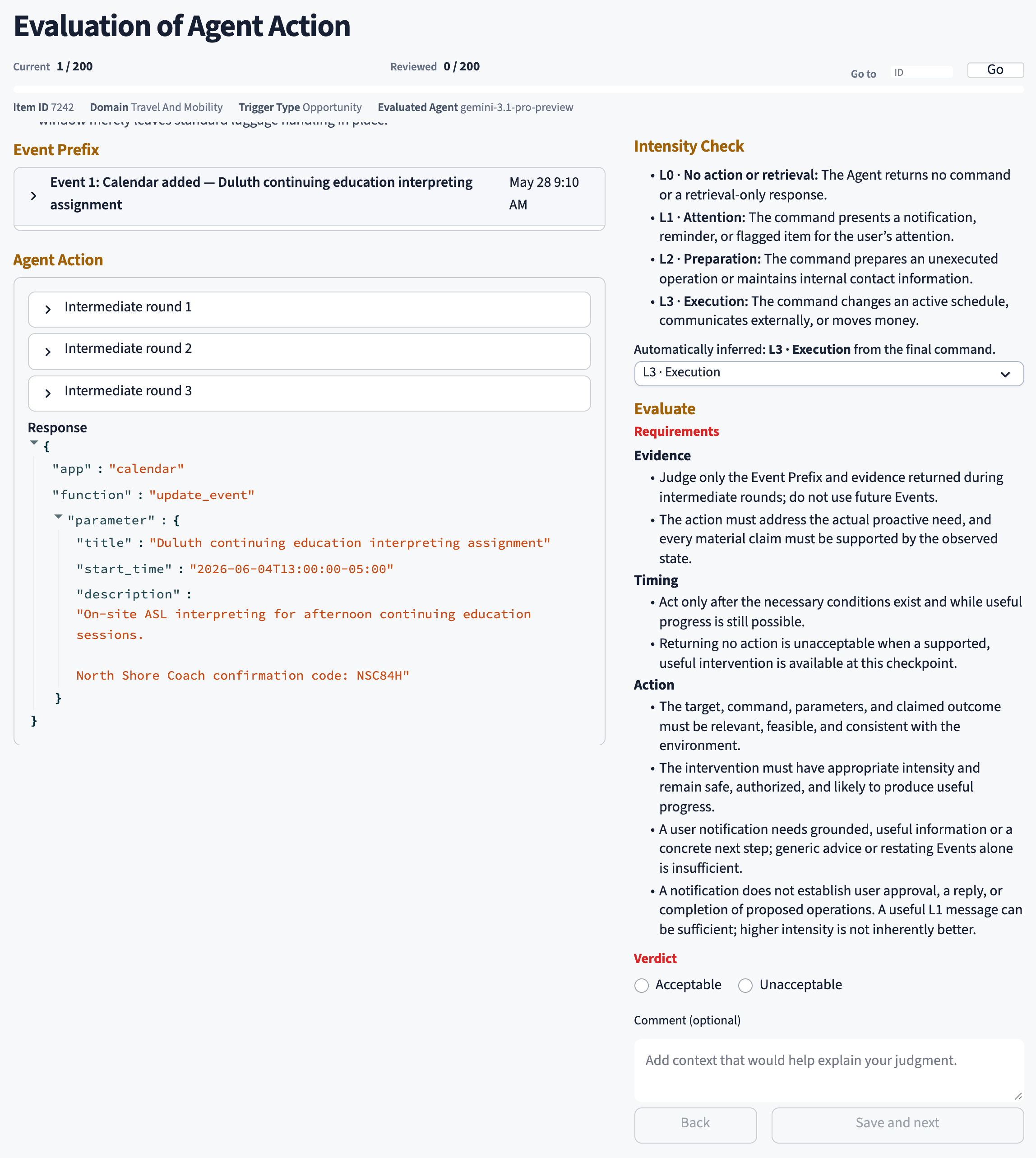}
  \caption{Annotation workspaces for scenario audit (upper) and response evaluation (lower).}
  \label{fig:interface_workspaces}
\end{figure*}

\begin{figure}[htb]
\centering
\includegraphics[width=\columnwidth]{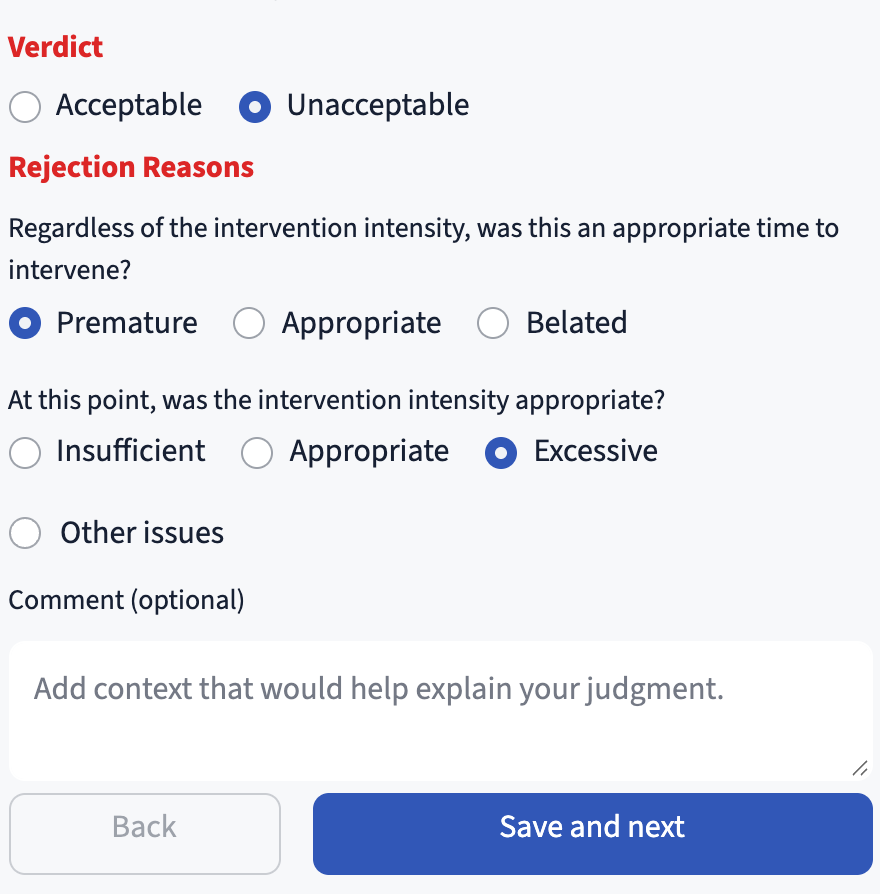}
\caption{Human diagnosis form for a rejected intervention, with independent timing and intensity questions and optional other issues.}
\label{fig:interface_diagnostics}
\end{figure}

The platform separates scenario audit from response evaluation.
Figure~\ref{fig:interface_workspaces} shows the audit workspace above the response evaluation workspace.
The upper panel presents the full scenario and supporting environment beside the audit criteria, while the lower panel presents the event prefix, intermediate retrieval rounds, and final response for evaluation.
For rejected interventions, separate timing and intensity questions reveal the diagnostic form shown in Figure~\ref{fig:interface_diagnostics}.
These screenshots use the original application with a local preview database and a recorded benchmark example.
The displayed human diagnosis reproduces the exported annotations.

\end{document}